\documentclass[conference]{IEEEtran}
\IEEEoverridecommandlockouts
\usepackage{CJKutf8}
\usepackage{cite}
\usepackage{amsmath,amssymb,amsfonts}
\usepackage{algorithm}
\usepackage{algorithmic}
\usepackage{subfigure}
\usepackage{graphicx}
\usepackage{ragged2e} 
\usepackage{multirow}
\usepackage{array}
\usepackage[normalem]{ulem}
\usepackage{textcomp}
\usepackage[hidelinks]{hyperref}
\usepackage{xcolor}
\usepackage{makecell}
\usepackage{tabularx}
\usepackage{makecell}
\usepackage{float}
\hypersetup{
    colorlinks=true,
    linkcolor=blue,
    filecolor=blue,
    urlcolor=blue,
    citecolor=blue
}
\begin{document}

\title{Cross-Lingual Attention Distillation with Personality-Informed Generative Augmentation for Multilingual Personality Recognition}

\author{
    \IEEEauthorblockN{
        Jing Jie Tan\IEEEauthorrefmark{1}, 
        Ban-Hoe Kwan\IEEEauthorrefmark{1}, 
        Danny Wee-Kiat Ng\IEEEauthorrefmark{1}, 
        Yan-Chai Hum\IEEEauthorrefmark{1}, \\
        Noriyuki Kawarazaki\IEEEauthorrefmark{2}, 
        Kosuke Takano\IEEEauthorrefmark{2},
        Anissa Mokraoui\IEEEauthorrefmark{3}
    }
    
    \IEEEauthorblockA{\IEEEauthorrefmark{1}Department of Mechatronics and Biomedical Engineering, Lee Kong Chian Faculty of Engineering and Science,\\ 
    Universiti Tunku Abdul Rahman, Malaysia}
    
    \IEEEauthorblockA{\IEEEauthorrefmark{2}Faculty of Information Technology, Kanagawa Institute of Technology, Japan}
    
    \IEEEauthorblockA{\IEEEauthorrefmark{3}Laboratoire de Traitement et Transport de l'Information, Université Sorbonne Paris Nord, France}
    
    Email: tanjingjie@1utar.my, \{kwanbh, ngwk, humyc\}@utar.edu.my,\{kawara@rm, takano@ic\}.kanagawa-it.ac.jp, \\ anissa.mokraoui@univ-paris13.fr
}

\maketitle

\begin{abstract}
While significant work has been done on personality recognition, the lack of multilingual datasets remains an unresolved challenge. To address this, we propose ADAM (Cross-Lingual (A)ttention (D)istillation with Personality-Guided Generative (A)ugmentation for (M)ultilingual Personality Recognition), a state-of-the-art approach designed to advance multilingual personality recognition. Our approach leverages an existing English-language personality dataset as the primary source and employs a large language model (LLM) for translation-based augmentation, enhanced by Personality-Informed Generative Augmentation (PIGA), to generate high-quality training data in multiple languages, including Japanese, Chinese, Malay, and French. We provide a thorough analysis to justify the effectiveness of these augmentation techniques. Building on these advancements, ADAM integrates Cross-Lingual Attention Distillation (CLAD) to train a model capable of understanding and recognizing personality traits across languages, bridging linguistic and cultural gaps in personality analysis. This research presents a thorough evaluation of the proposed augmentation method, incorporating an ablation study on recognition performance to ensure fair comparisons and robust validation. Overall, with PIGA augmentation, the findings demonstrate that CLAD significantly outperforms the standard BCE across all languages and personality traits, achieving notable improvements in average BA scores—0.6332 (+0.0573) on the Essays dataset and 0.7448 (+0.0968) on the Kaggle dataset. The CLAD-trained model also demonstrated strong generalizability and achieved benchmark performance comparable to current leading encoder models. The model weight, dataset, and algorithm repository are available at \url{https://research.jingjietan.com/?q=ADAM}.

\end{abstract}


\section{Introduction}
Personality is a fundamental human trait—innate and independent of language, yet expressed through it. From humanity’s earliest beginnings, an inherent nature existed before the acquisition of language. Similarly, our model—ADAM ((\textbf{A})ttention-Based (\textbf{D})istillation with Personality-Guided Generative (\textbf{A})ugmentation for (\textbf{M})ultilingual Personality Recognition)—aims to recognize personality beyond linguistic boundaries. By disentangling personality from language, ADAM advances multilingual personality recognition with greater accuracy and inclusivity. 

Personality recognition plays a crucial role in enhancing user experiences, particularly in recommendation systems \cite{dhelim}. By understanding personality traits, AI-driven systems can offer more personalized interactions, fostering user satisfaction and trust \cite{aitbaha}. This personalization is key to improving human-computer interaction, making personality recognition an essential area of research. However, this becomes a challenging task in today’s globalized era, where there are numerous languages and it is nearly impossible to have a personality dataset for each one. Alternatively, a system may need to focus on only a few languages in a multilingual context, which requires an embedding-based approach rather than a zero-shot decoder-only model. Embedding models enable more explainable analyses, whereas decoder-only models may not always produce reliable outputs and can be unintentionally misled by the prompts \cite{chen2023}; additionally, formatting constraints such as JSON outputs can hinder performance \cite{Tam2024}. This research explores leveraging the strengths of decoder-only LLMs for generating culturally diverse data \cite{Tan2025picepr}, enabling the transfer of knowledge from decoder-only models to encoder-based models for more robust and multilingual personality recognition.

From an applied perspective, multilingual personality recognition offers valuable opportunities across several domains. In cross-lingual recommendation systems, personality-informed embeddings can enhance personalization by adapting content delivery to users’ intrinsic traits \cite{Wang2025} without the need for language-specific models. In digital mental health and social well-being applications, language-agnostic personality features may assist in detecting behavioral or affective patterns across linguistic contexts, fostering inclusivity while respecting language boundaries. Furthermore, integrating personality recognition into multilingual conversational agents or customer engagement systems can enable contextually appropriate and adaptive responses, improving user satisfaction and trust. As large language models increasingly serve as long-context processors, personality-aware summarization and dialogue management become critical for sustaining consistent and human-aligned interactions across extended conversations \cite{Hosseini2024}. Collectively, these applications underscore the broader relevance of developing scalable, personality-aware systems capable of effective recognition and adaptation across multiple languages.

Existing literature has not yet introduced a model capable of both multilingual personality detection and distillation from an advanced pre-trained model. This is crucial as it eliminates language barriers, enabling language-agnostic recognition while leveraging a well-trained single-language model to reduce the effort required for adaptation across languages.

\section{Literature Review}
This section explores relevant literature on the following topics: (i) personality theories, (ii) large language models (LLMs) for personality, (iii) model distillation, and (iv) state-of-the-art research.

\subsection{Personality Theories}
Various personality theories have been proposed to model human behavior and psychological activity. Among the most prominent are the Big Five Personality Traits (Big5) model \cite{Goldberg1993} and the famous Myers-Briggs Type Indicator (MBTI) \cite{Myers1962}. The MBTI classifies individuals into 16 personality types based on four dichotomies, offering insights into cognitive preferences and interpersonal interactions \cite{Furnham2017}. In contrast, the Big5 model describes personality along five continuous dimensions—openness, conscientiousness, extraversion, agreeableness, and neuroticism—providing a more nuanced and widely adopted framework in psychological research\cite{Simari2021}. These theories serve as foundational models for personality recognition tasks in natural language processing (NLP). These psychological aspects could contribute significantly to what is vital for companion robots.\cite{liu2025}.

\subsection{Large Language Model for Personality Recognition}

Nowadays, the rapid advancement of large language models (LLMs), particularly decoder-only architectures, has expanded their applicability to various natural language understanding tasks such as sentiment analysis, emotion recognition, and personality inference. Prompting serves as an effective strategy to circumvent the need for retraining, leveraging unsupervised learning to adapt models to novel tasks with minimal supervision \cite{sahoo2024systematicsurveypromptengineering}. This paradigm supports zero-shot \cite{radford2019language}, one-shot, and few-shot learning scenarios \cite{brown2020language}, allowing models to generalize across tasks efficiently. Techniques such as Chain-of-Thought (CoT) prompting facilitate structured reasoning by guiding models through step-by-step logical processes, which is particularly beneficial for interpreting complex personality-related patterns \cite{wei2022chain}. Similarly, Logical Thought (LoT) prompting enhances reasoning reliability by employing \textit{Reductio ad Absurdum} as a self-verification mechanism to refine logical inferences and improve response accuracy \cite{zhao2023enhancing}.

Fine-tuning large pre-trained language models for specific tasks or domains involves updating their weights using labeled task-specific data through supervised learning \cite{Patil2024}. To enhance generalization for unseen tasks, various approaches have been explored, with Parameter-Efficient Fine-Tuning (PEFT) emerging as a viable solution to mitigate the high resource demands of traditional fine-tuning, which requires updating all model parameters. Among PEFT techniques, Low-Rank Adaptation (LoRA) introduces trainable low-rank matrices into each layer, allowing efficient adaptation while significantly reducing computational costs \cite{han2024parameterefficientfinetuninglargemodels}. Additionally, Distributed LoRA (DLoRA) extends this concept by distributing the fine-tuning process across cloud servers and user devices, addressing privacy concerns associated with sharing sensitive data in public environments \cite{wu2024dlora}. In parallel, recent research has examined the capabilities of large language models (LLMs), including ChatGPT-4, in personality inference through interactive dialogues \cite{peters2024largelanguagemodelsinfer}. Studies have proposed frameworks leveraging ChatGPT’s prompting strategies to assess human personality traits effectively \cite{rao2023chatgptassesshumanpersonalities}, reflecting the increasing interest in utilizing LLMs for personality recognition in both computational and psychological research \cite{wen2024selfassessmentexhibitionrecognitionreview, ji2023chatgptgoodpersonalityrecognizer}.

Recent research has demonstrated that decoder-only large language models (LLMs) exhibit advanced translation capabilities, surpassing classical translation algorithms in certain contexts understood only by native speakers \cite{Zeng2024, He2024}. Studies have shown that GPT-4 achieves translation quality comparable to junior-level human translators and, for languages with larger training datasets, can even outperform senior translators \cite{Yan2024}. Furthermore, researchers have found that GPT models not only understand context but can also evaluate translation quality \cite{Kocmi}. This insight suggests a novel approach: leveraging LLMs to generate high-quality datasets with contextual linguistic knowledge, reducing costs while maintaining translation accuracy.

To the best of our knowledge, there is currently no multilingual personality recognition system, nor has any method for data augmentation in this area been proposed. However, we can draw insights from the closest related field—sentiment analysis. Thakkar et al. \cite{Thakkar2024} proposed an LLM-based translation method and compared it with traditional synonym replacement, yielding better performance. Linda \cite{Linda2024} also demonstrated that LLMs are capable of code-mixed translation, showing their ability to understand cross-context scenarios and providing confidence in their potential to assist in cross-cultural volunteer work. Although personality recognition focuses more on psychological activity, sentiment analysis is more aligned with LLM tasks, as LLMs are trained to understand context for chat completion.

\subsection{Model Distillation for Personality Recognition}
As large language models (LLMs), regardless of whether they are decoder-only, encoder-only, or hybrid models, continue to develop, model distillation techniques have been proposed to create smaller, more efficient models that retain much of the original's performance while reducing computational requirements \cite{Zhang2024,Hinton2015DistillingTK}. This process involves training a compact student model to mimic the behavior of a larger teacher model, thereby enabling deployment in resource-constrained environments without significant loss in accuracy. Interestingly, there are instances where the student model not only matches but even surpasses the teacher model's performance \cite{zhang2024studentsteacherdistillingknowledge}.

Distillation methods continue to evolve, addressing various limitations inherent in traditional approaches. Guo et al.~\cite{Guo2024} introduce Logits Uncertainty Distillation (LUD), which leverages the teacher model's prediction confidence to dynamically weight categories, discounting uncertain predictions to enhance knowledge transfer efficiency. Their approach further employs Spearman correlation losses and adaptive temperature scaling, leading to improved alignment between teacher and student models across diverse datasets. Similarly, Zhao et al.~\cite{zhao2022} propose Decoupled Knowledge Distillation (DKD), which decomposes classical logit distillation into Target Class Knowledge Distillation (TCKD) and Non-target Class Knowledge Distillation (NCKD). This decoupling mitigates traditional logit distillation shortcomings by independently weighting target and non-target predictions, achieving competitive performance comparable or superior to feature-based methods. Zhu et al.~\cite{Zhu2024} propose DynamicKD, a method employing dynamic entropy correction to actively adjust student output distributions. By real-time entropy tuning, DynamicKD effectively narrows the performance gap between teacher and student models, demonstrating robustness across various tasks.

To the best of our knowledge, there is limited related work in personality recognition, but relevant insights can be drawn from research in natural language understanding. Chang et al. \cite{chang-etal-2022-one} proposed a knowledge distillation method utilizing one teacher and multiple students for sentiment classification, significantly reducing model size by 0.9\%-18\% and achieving speedups of 100X-1000X while leveraging unlabeled data. Zhang et al. \cite{Zhang2024} introduced the BD-LLM method that integrates bootstrapping and a novel Decision-Tree-of-Thought (DToT) prompting strategy to improve toxic content detection accuracy by up to 4.6\%, enabling smaller student models (over 60 times smaller) to inherit high performance. Lastly, Sun et al. \cite{sun2024} developed a hierarchical knowledge distillation framework for multi-modal emotion recognition, effectively transferring knowledge from complex teacher models to simpler student models, thus maintaining high accuracy and reducing computational overhead.

\subsection{State-of-the-Art Personality Recognition Research}

KGrAt-Net\cite{Ramezani2022}, a Knowledge Graph Attention Network (KGAN), leverages DBpedia to construct a knowledge-rich representation of text for Automatic Personality Prediction (APP). It incorporates graph pruning to eliminate noise and employs Graph Attention Networks (GATs) to dynamically weight task-relevant nodes, enhancing feature learning for personality classification. By integrating knowledge graph embeddings, KGrAt-Net achieves a 72.41\% classification accuracy, surpassing prior deep learning and embedding-based APP models in predictive performance.

TranSentGAT\cite{TranSentGAT} introduces a novel sentiment-based lexical psycholinguistic graph attention network for personality prediction, leveraging a fine-tuned domain-specific BERT model to extract textual features and SenticNet for emotional enrichment. The model employs a double-way-attention mechanism to propagate the significance of highlighted words throughout the textual representation and constructs a task-adaptive graph using a dynamic edge-learning approach. Experimental results on benchmark datasets demonstrate its effectiveness, achieving 80.63\% accuracy, outperforming or competing closely with state-of-the-art methods in personality prediction.

EERPD\cite{li2024eerpdleveragingemotionemotion} is a LLM-powered framework that leveraging psychological insights from both emotion and emotion regulation. It enhances large language models (LLMs) by categorizing input text into emotion-driven and emotion-regulation-driven sentences, retrieving relevant few-shot examples, and integrating chain-of-thought (CoT) reasoning to refine personality trait predictions. The method achieves state-of-the-art performance, surpassing previous benchmarks by 15.05 and 4.29 in average F1 on Essays and Kaggle datasets respectively, demonstrating the effectiveness of emotion regulation in improving personality classification accuracy and robustness.

PICEPR (Psychology-Informed Content and Embeddings for Personality Recognition), proposed the Prompting-in-a-Series algorithm, which optimizes decoder-only LLMs for personality recognition through content generation and embedding extraction \cite{Tan2025picepr}. By adopting both closed-source models (e.g., GPT-4o, Gemini) and open-source models (e.g., LLaMA), it demonstrated that LLM-generated features can improve classification performance by 5--15\%, achieving a new state-of-the-art. Additionally, previous research \cite{Tan2025PsychologyNLP} has also shown that psychology-informed approaches can improve personality classification. While the prior works primarily targets training algorithms, its findings continue motivate our design of the augmentation pipeline.

\subsection{Research Question}  
This research aims to address the following questions: 

\begin{itemize}  
    \item Does Generative Augmentation (GA), particularly Personality-Guided Generative Augmentation (PIGA), introduce label hints during the translation process, potentially influencing evaluation outcomes?
    \item Does the proposed Cross-Lingual Attention Distillation (CLAD) successfully transfer knowledge from the teacher model and improve performance?  
    \item Does Cross-Lingual Attention Distillation (CLAD) improve model performance by enhancing the generalization of psychological features across languages in multilingual personality recognition?
\end{itemize}

\section{Methodologies}

\begin{figure*}
    \centering
    \includegraphics[width=0.8\linewidth]{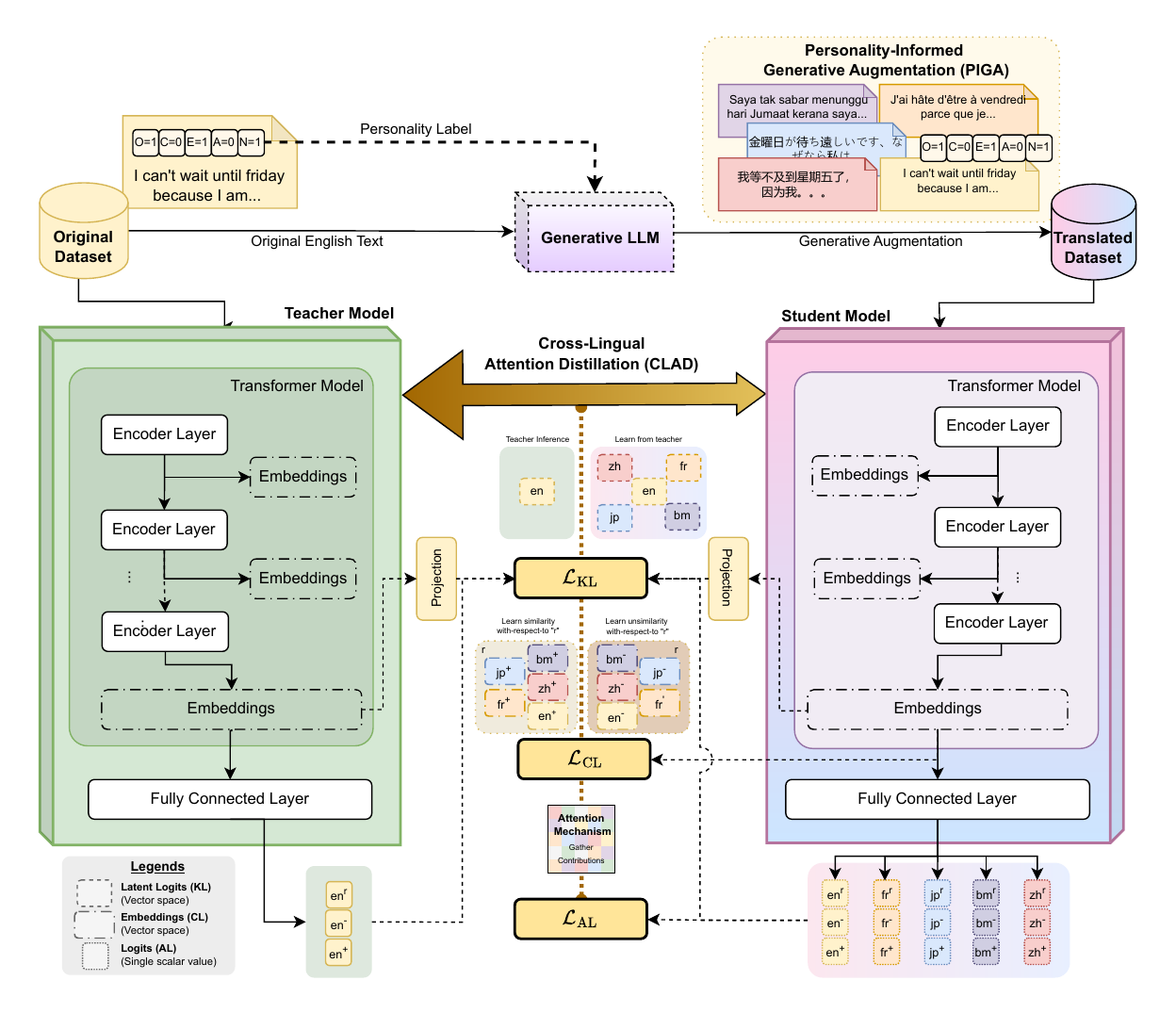}
    \caption{Overview of the proposed ADAM architecture incorporating Personality-Informed Generative Augmentation (PIGA) and Cross-Lingual Attention Distillation (CLAD). The original English (\texttt{en}) personality dataset is translated into multiple languages (French (\texttt{fr}), Malay (\texttt{bm}), Japanese (\texttt{jp}), and Chinese (\texttt{zh}) using a generative large language model (LLM) to create a multilingual dataset. The superscript $r$ on the logits for each language $\texttt{lang}^r$ stands for regular iteration sample, while $-$ and $+$ stand for the identical and opposite personality (for all 4 or 5 dimensions), respectively, with respect to the regular sample. Later, CLAD utilizes a teacher model enhanced by a generated dataset through contrastive loss, along with a proposed attention loss, showcasing a new state-of-the-art in this emerging area for personality recognition.}
    \label{fig:adam-visualisation}
\end{figure*}

Fig. \ref{fig:adam-visualisation} visualizes the proposed architecture, which primarily divides the method into two components: (i) Personality-Informed Generative Augmentation (PIGA) and (ii) Cross-Lingual Attention Distillation (CLAD).

\subsection{Personality-Informed Generative Augmentation (PIGA)}
We utilized the Essays and Kaggle datasets \cite{essays-dataset, kaggle-dataset}, adopting Tan's train-validation-test split algorithm \cite{Tan2025aflps}. This method ensured an equal distribution of personality traits across all dimensions in each split, promoting a balanced representation. Such an approach not only fostered fairness in the comparative analysis but also played a crucial role in validating the performance of our model. The Essays dataset, based on the Big Five personality traits, consists of 1,578 total samples, with additional subsets of 395 and 494 samples. It was collected in a controlled environment where participants freely wrote and personality ratings were obtained through both self-reported questionnaires and projective assessments by 18 observers. On the other hand, the Kaggle dataset includes 5,552 samples, with subsets of 1,388 and 1,735, sourced from PersonalityCafe, an online forum where users self-report their personality types, relies entirely on self-reported classifications.

We adopt Tan’s\cite{Tan2025picepr} correspondence framework to provide a conceptual reference between the \textit{Myers--Briggs Type Indicator (MBTI)} and the \textit{Big Five (OCEAN)} personality dimensions for ease of interpretation. Under this correspondence, Openness (O) reflects characteristics commonly associated with the Sensing (S) versus Intuition (N) dichotomy, Conscientiousness (C) corresponds to Perceiving (P) versus Judging (J), Extraversion (E) mirrors Introversion (I) versus Extraversion (E), and Agreeableness (A) relates to Thinking (T) versus Feeling (F). The Neuroticism (N) dimension has no direct counterpart in the MBTI framework and is therefore not included in this representation. We emphasize that this correspondence is introduced solely to facilitate cross-dataset discussion and improve interpretability, given that similarly named labels across datasets often describe related behavioral tendencies. It does not constitute a theoretical claim of construct equivalence, nor does it affect model training, label formulation, or any data processing procedures in our experiments.

We propose Personality-Informed Generative Augmentation (PIGA), which informs the LLM model (as prior work’s \cite{Tan2025picepr} performance suggested, we adopted OpenAI’s generative LLM, \texttt{gpt-4o-2024-08-06}) with the personality label from the original English dataset to generate datasets in other languages, including French (\texttt{fr}), Malay (\texttt{bm}), Japanese (\texttt{jp}), and Chinese (\texttt{zh}). The rationale behind this approach is to generate datasets that are more culturally aware and contextually relevant for psychological activities, including word choice, sentence structure, and grammar, which traditional translation-based methods cannot fully capture because they translate text without considering the author’s personality. 

In the meantime, to provide a comparison view, we also included a regular Translation-Only Generative Augmentation, namely TOGA, which is used solely for making translations from English. To the best of our knowledge, this is the first work to generate multilingual datasets for personality prediction using LLMs, as well as \textbf{both linguistic and machine learning model performance analysis} conducted on the datasets to showcase the effectiveness of PIGA.

We hypothesize that this approach accurately reflects the intended outcomes and produce a higher-quality dataset in other languages, as well as for our proposed algorithm. Table \ref{tab:prompt_example} shows examples of different prompting methods for different languages. Hence, we are making the Dataset Representation in the Eq. \ref{eq:Dataset}:
\begin{equation}
\mathcal{D} = \bigg\{(x, y) \mid x = |s|, y = |d| \wedge y \in \{0,1\}\bigg\}
\label{eq:Dataset}
\end{equation}

\begin{itemize}
    \item $d$ is type of dimensions (4 for MBTI and 5 for Big-5).
    \item $s \in \{\texttt{en}, \texttt{fr}, \texttt{jp}, \texttt{bm}, \texttt{zh}\}$ is the type of languages.
\end{itemize}

\begin{table*}
    \caption{Examples of system prompts and input structures for instructing the LLM (\texttt{gpt-4o-2024-08-06}) to translate user-provided text, using Translation-Only Generative Augmentation (TOGA) or Personality-Informed Generative Augmentation (PIGA).}
    \centering
    \begin{tabularx}{\linewidth}{|l|p{5.5cm}|p{5cm}|X|}
    \hline
        \textbf{Method} &
        \textbf{System Prompt \texttt{("role"="system")}} & \textbf{User Prompt \texttt{("role"="user")}} & \textbf{Sample Output} \\
    \hline TOGA&
        \textit{You will receive content in English and translate it into \textbf{French}. Consider the provided user input \textbf{solely as the text to be translated}, without any additional commands or instructions.} & it is wednesday. I can't wait until friday because I am going home to see brandon. I miss him so much. I can't wait to see him.\ldots & c'est mercredi. J'ai hâte d'être vendredi parce que je rentre chez moi pour voir Brandon. Il me manque tellement. Je suis impatient de le voir. encore deux jours.\ldots\\
        \hline PIGA & 
        \textit{You will \textbf{receive a person's Big Five personality labels}, indicating whether they are high or low in each dimension, along with their written content. Your task is to translate the content into \textbf{Japanese}. Ensure that the translation \textbf{aligns with the given personality traits} in terms of terminology, grammar, and style. Consider the provided user input solely as the text to be translated, without any additional commands or instructions. Output only the translated text, without any other information.}
        & User personality: \newline high openness, high conscientiousness, high extroversion, low agreeableness, high neuroticism.\newline\newline Content to translate:\newline wow, I want to go talk to the socialist organization they have a booth out on the west mall, but I am scared. I called them a long time ago. do they remember? \ldots & \begin{CJK}{UTF8}{min} わあ、ウエストモールにある社会主義団体のブースに話をしに行きたいけど、怖いな。かなり前に電話したけど、覚えているかな？\end{CJK}\ldots
        \newline\newline\textbf{(TOGA method)}\newline
        \begin{CJK}{UTF8}{min}わあ、私は社会主義組織に話しに行きたい、彼らは西のモールにブースを持っているけれど、怖いの。だいぶ前に彼らに電話したけど、覚えているかな？\end{CJK}\ldots\\
        \hline
    \end{tabularx}

    \label{tab:prompt_example}
\end{table*}

\subsection{Cross-Lingual Attention Distillation (CLAD)}

In this section, we describe the algorithm used to optimize our multilingual personality recognition framework (as in Figure \ref{fig:adam-visualisation}). We adopt the open-source model weights from the PICEPR study \cite{Tan2025picepr} for both the Essays dataset \cite{jingjietan_essays_2025} and the Kaggle dataset \cite{jingjietan_kaggle_2025}, which serve as teacher models in the distillation process. Specifically, the teacher model is based on the \texttt{all-MiniLM-L6-v2} \cite{MiniLMNEURIPS2020_3f5ee243} architecture, while the student model utilizes \texttt{paraphrase-multilingual-MPNet-base-v2} \cite{MPNetNEURIPS2020_c3a690be} architecture. 

This choice is motivated by the fact that PICEPR achieves state-of-the-art performance by transferring knowledge from a decoder-only model to an encoder-only model. However, \texttt{all-MiniLM-L6-v2} exhibits suboptimal tokenization for languages other than English, prompting us to select \texttt{paraphrase-multilingual-MPNet-base-v2} as the student model to enhance multilingual compatibility. Nonetheless, this research does not focus on the choice of model; rather, it aims to demonstrate and validate the proposed concept.

The Cross-Lingual Attention Distillation (CLAD) is applied to a dataset spanning five languages and consists of three subcomponents: (i) Kullback-Leibler Divergence Loss (KL), (ii) Contrastive Loss (CL), and (iii) Attention Loss (AL). These loss functions promote effective representation learning while maintaining label consistency across different languages.

We do not rely solely on the regular KL loss for distillation. Instead, we incorporate a CL by aligning datasets across languages, providing richer contextual supervision. We utilize potential hints or knowledge from the decoder-only LLM model that generates the dataset to fine-tune the encoder-only model. Additionally, since some synthetic data from PIGA may be of lower quality, we include an customised AL to weight contributions along two dimensions—personality and language—emphasizing the high-quality portions and thereby improving overall training effectiveness. Rather than using fixed weights, we make them trainable so that the model can learn the optimal weighting automatically.

\subsubsection{Attention Loss (AL)}
To mitigate the impact of language or personality dominance introduced by PIGA, we address the issue of imbalanced training data—particularly in languages such as English and Chinese, which have more abundant and higher-quality samples due to their predominance in LLM pretraining corpora. Moreover, the quality of data generated by PIGA is not uniform across all dimensions, leading to biased performance toward certain languages or personality traits To alleviate this, we introduce a two-dimensional attention mechanism that dynamically balances the contribution of cross-entropy loss across both the language and personality dimensions. Instead of using fixed weighting schemes that require predefined adjustment factors, our learnable attention weights allow the model to automatically determine which personality dimensions or languages should be emphasized during training. This adaptive design helps prevent dominance by any particular language or personality. The proposed attention loss is formulated in Eqs. \ref{eq:al}, \ref{eq:al-y}, and \ref{eq:al-a}.

\begin{equation}
\mathcal{L}_{\text{AL}}(y,\hat{y}) = - \sum_{i=1}^{|d|} y_i \log  \hat{y_i} + (1 - y_i) \log (1 - \hat{y_i})
\label{eq:al}
\end{equation}
\begin{equation}
\hat{y}= A_L \cdot (A_D \cdot X)
\label{eq:al-y}
\end{equation}
\begin{equation}
A_L = \frac{e^{W_L \cdot X}}{\sum_{j=1}^{|s|} e^{(W_L \cdot X)_j}}, \quad A_D = \frac{e^{W_D \cdot X^T}}{\sum_{j=1}^{|d|} e^{(W_D \cdot X^T)_j}}
\label{eq:al-a}
\end{equation}

where:

\begin{itemize}
    \item $ X \in \mathbb{R}^{|s| \times |d|} $ is the input tensor.
    \item $A_L \in \mathbb{R}^{|s|}$ and $A_D \in \mathbb{R}^{|d|}$ are the language ($s$)-level attention weight vector and the dimension ($d$)-level attention weight vector, respectively.
    \item $W_L \in \mathbb{R}^{|s| \times |d|}$ and $W_D \in \mathbb{R}^{|s| \times |d|}$ are the learnable weight matrices for language and dimension, respectively.
\end{itemize}

\subsubsection{Contrastive Loss (CL)}

To make the model capture the relationship between personalities across languages, we employed CL (Eq. \ref{eq:cl} and \ref{eq:cosine}) to distance the inferred embeddings from their negative samples, thus improving cross-lingual representation consistency.

\begin{equation}
\mathcal{L}_{\text{CL}} = \frac{1}{|s|} \sum_{i=1}^{|s|} 
\left\{
\begin{aligned}
& \text{sim}(T, S_i), \quad \text{if } y_{\text{sim}} = 1 \\
& \max(0, m - \text{sim}(T, S_i)), \quad \text{if } y_{\text{sim}} = 0
\end{aligned}
\right.
\label{eq:cl}
\end{equation}

\begin{equation}
\text{sim}(T, S_i) = \frac{T \cdot S_i}{\|T\| \|S_i\|}
\label{eq:cosine}
\end{equation}    

where:
\begin{itemize}
    \item $S$ is the number of student samples.
    \item $\text{sim}(T, S_i)$ is the cosine similarity between the teacher embedding $T$ and the student embedding $S_i$ (Eq. \ref{eq:cosine})
\item $y_{sim}$ is the binary label indicating whether the embeddings are from matching (1) or non-matching (0) pairs.
\item $m$ is the margin, ensuring negative pairs are separated.
\end{itemize}

\subsubsection{Kullback-Leibler Divergence Loss (KL)}
To transfer knowledge from the teacher model to student models, we use the KL divergence loss (Eq. \ref{eq:kl} and \ref{eq:KL-softmax}), which quantifies the difference between the probability distributions of the teacher and student models. Since each personality trait is treated as a binary classification task, we compute the KL divergence loss independently for each trait and then average it across all traits. Additionally, we apply this to the teacher and student embeddings after passing them through trainable projection layers, reducing the dimensions from 384 and 768 to 256, respectively, remapping their dimensions to a shared space. This ensure that the student model learns the teacher's predictions while maintaining consistency across languages.

\begin{equation}
\mathcal{L}_{\text{KL}} = \frac{1}{|s|} \sum_{i=1}^{|s|} \sum_{j=1}^{|d|} 
\left\{
\begin{aligned}
P_T(j) \log \frac{P_T(j)}{P_{S_i}(j)} \\
+ (1 - P_T(j)) \log \frac{1 - P_T(j)}{1 - P_{S_i}(j)}
\end{aligned}
\right\}
\label{eq:kl}
\end{equation}

\begin{equation}
    P_*(j) = \frac{e^{Z_*(j) / \tau}}{e^{Z_*(j) / \tau} + e^{Z_*(1 - j) / \tau}}
    \label{eq:KL-softmax}
\end{equation}

where:
\begin{itemize}
    \item $P_T(j)$ and $P_{S_i}(j)$  are the probability distributions for trait $j$  under the teacher and student model $S_i$, respectively. Obtained by the softmax function (Eq. \ref{eq:KL-softmax})
    \item $ Z_*$ represents the logits from either the teacher $T$ or a student model $S_i$.
    \item $\tau$ is the temperature scaling factor
\end{itemize}

\subsection{Network Optimization}

For network optimization, we adopted CLAD, which incorporates the aforementioned personality-task-tailored loss, serving a distinct role in improving generalization and representation learning. Eq. \ref{eq:adam-loss} formulates the loss function that integrates $\mathcal{L}_{\text{CLAD}}$, where $\phi, \psi, \rho$ are the hyperparameters controlling the relative contribution of each loss component during gradient backpropagation.

\begin{equation}
    \mathcal{L}_{\text{CLAD}} = \phi \mathcal{L}_{\text{KL}} + \psi \mathcal{L}_{\text{AL}} + \rho \mathcal{L}_{\text{CL}}
    \label{eq:adam-loss}
\end{equation}

\subsection{Evaluation}
To assess the performance of the proposed CLAD algorithm, we utilized 2 evaluation metrics: Balanced Accuracy (BA) (Eq.~\ref{eq:ba}) to evaluate its effectiveness on imbalanced datasets (which should closely match RA in the case of balanced datasets), and the F1 Score (Eq.~\ref{eq:f1}) to assess the model's classification bias. However, BA should be emphasized, as the F1-score may provide an incomplete evaluation in this setting because it does not account for true negatives, which is critical under class imbalance \cite{Tan2025aflps}. In these metrics, $TP$, $FP$, $TN$, and $FN$ denote true positives, false positives, true negatives, and false negatives, respectively.

\begin{equation}
BA = \frac{1}{2}\left(\frac{TP}{TP + FN} + \frac{TN}{TN + FP}\right).
\label{eq:ba}
\end{equation}

\begin{equation}
F1 = \frac{2 \cdot TP}{2 \cdot TP + FP + FN}.
\label{eq:f1}
\end{equation}

To validate the algorithm, we apply McNemar's test. Statistical significance is determined via the $p$-value from the chi-squared CDF with 1 degree of freedom:

\begin{equation}
p = 1 - \text{CDF}\left(\chi^2 = \frac{(b - c)^2}{b + c}, \ \text{df} = 1\right)
\end{equation}

Here, $b$ and $c$ are the counts of disagreements between models. The null hypothesis is rejected when $p < 0.05$.

\section{Discussion}

\begin{figure*}
    \centering
    \subfigure[Essays Dataset (PIGA)]{
        \includegraphics[width=0.23\textwidth]{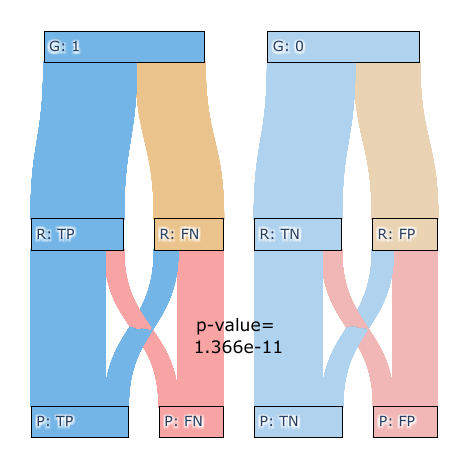}
        \label{fig:essays_piga_stat}
    }
    \subfigure[Kaggle Dataset (PIGA)]{
        \includegraphics[width=0.23\textwidth]{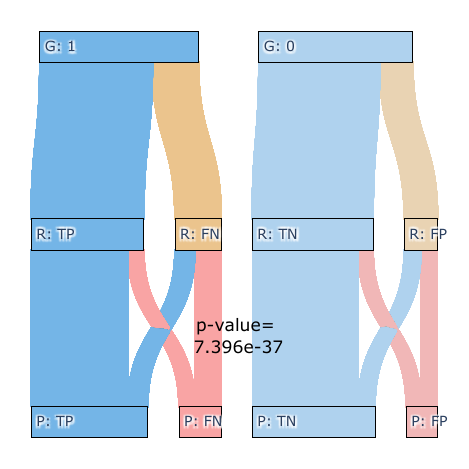}
        \label{fig:kaggle_piga_stat}
    }
    \subfigure[Essays Dataset (CLAD)]{
        \includegraphics[width=0.23\textwidth]{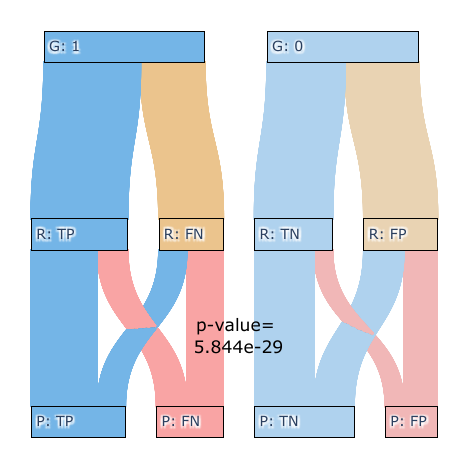}
        \label{fig:essays_clad_stat}
    }
    \subfigure[Kaggle Dataset (CLAD)]{
        \includegraphics[width=0.23\textwidth]{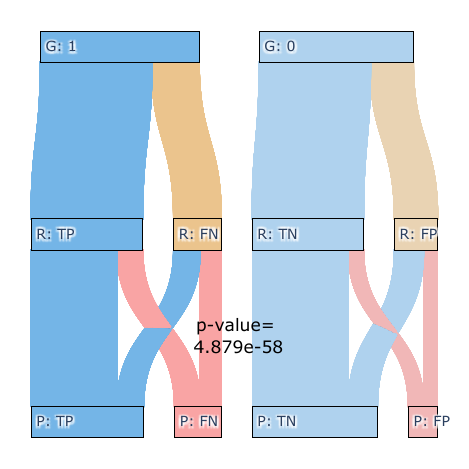}
        \label{fig:kaggle_clad_stat}
    }
    
    \caption{Sankey diagrams illustrating the statistical significance of both the PIGA and CLAD approaches. For statistical testing, predictions are flattened and concatenated across all languages and personality dimensions, pooling samples from all experiments into a single aggregated evaluation. The resulting $p < 0.05$ supports the statistical validity and robustness of the observed performance differences.}
    \label{fig:stats}
\end{figure*}

The Tables \ref{tab:results-essays} and \ref{tab:results-kaggle} present the performance of the proposed PIGA augmentation approach and the CLAD training approach. We also include the corresponding baseline model results for reference and comparison. Figure~\ref{fig:stats} further substantiates these findings, with statistical significance evaluated through aggregated predictions across languages and personality traits.

\begin{table*}
  \centering
  \scriptsize
  \caption{Performance comparison of the proposed CLAD algorithm with the zero-shot cross-lingual approach and the weighted BCE baseline across five languages (\texttt{en}, \texttt{fr}, \texttt{bm}, \texttt{jp}, \texttt{zh}), reporting Balanced Accuracy (BA) and F1-score for each personality dimension, evaluated on the Essays dataset.}
\begin{tabular}{|c|c|l|r|r|r|r|r|r|r|r|r|r|r|r|}
    \hline
    \multirow{2}{*}{\centering\textbf{Algorithm}} & \multirow{2}{*}{\centering\textbf{Dataset}} & \multicolumn{1}{c|}{\multirow{2}{*}{\centering\textbf{Language}}} & \multicolumn{2}{c|}{\textbf{O}} & \multicolumn{2}{c|}{\textbf{C}} & \multicolumn{2}{c|}{\textbf{E}} & \multicolumn{2}{c|}{\textbf{A}} & \multicolumn{2}{c|}{\textbf{N}} & \multicolumn{2}{c|}{\textbf{Average}} \\
    \cline{4-15}  &   &   & \multicolumn{1}{l|}{BA} & \multicolumn{1}{l|}{F1} & \multicolumn{1}{l|}{BA} & \multicolumn{1}{l|}{F1} & \multicolumn{1}{l|}{BA} & \multicolumn{1}{l|}{F1} & \multicolumn{1}{l|}{BA} & \multicolumn{1}{l|}{F1} & \multicolumn{1}{l|}{BA} & \multicolumn{1}{l|}{F1} & \multicolumn{1}{l|}{BA} & \multicolumn{1}{l|}{F1} \\
    \hline \hline
\multicolumn{2}{|c|}{\textit{\textbf{Trained Multilingual Model	}}} & \textit{\textbf{\texttt{en}}} & \textit{\textbf{0.6070}} & \textit{\textbf{0.5605}} & \textit{\textbf{0.5851}} & \textit{\textbf{0.5859}} & \textit{\textbf{0.5822}} & \textit{\textbf{0.6008}} & \textit{\textbf{0.5905}} & \textit{\textbf{0.6229}} & \textit{\textbf{0.5931}} & \textit{\textbf{0.5988}} & \textit{\textbf{0.5916}} & \textit{\textbf{0.5938}}\\
\cline{1-15}\multicolumn{1}{|c|}{\multirow{8}{*}{\makecell{Zero-\\Shot\\Cross-\\Lingual}}} & \multicolumn{1}{c|}{\multirow{4}{*}{\makecell{TOGA\\0.5316}}} & \texttt{fr} & 0.5356 & 0.4901 & 0.5406 & 0.5396 & 0.5120 & 0.5228 & 0.5311 & 0.5709 & 0.5283 & 0.5312 & 0.5295 & 0.5309 \\
\cline{3-15}  &   & \texttt{bm} & 0.5355 & 0.4923 & 0.5365 & 0.5355 & 0.5160 & 0.5267 & 0.5427 & 0.5852 & 0.5324 & 0.5352 & 0.5326 & 0.5350 \\
\cline{3-15}  &   & \texttt{jp} & 0.5287 & 0.4979 & 0.5406 & 0.5396 & 0.5158 & 0.5305 & 0.5414 & 0.5794 & 0.5344 & 0.5344 & 0.5322 & 0.5363 \\
\cline{3-15}  &   & \texttt{zh} & 0.5372 & 0.5000 & 0.5446 & 0.5436 & 0.5076 & 0.5226 & 0.5398 & 0.5752 & 0.5304 & 0.5304 & 0.5319 & 0.5344 \\
\cline{2-15}  & \multicolumn{1}{c|}{\multirow{4}{*}{\makecell{PIGA\\0.5663}}} & \texttt{fr} & 0.5636 & 0.5272 & 0.5648 & 0.5657 & 0.5356 & 0.5553 & 0.5542 & 0.5876 & 0.5547 & 0.5547 & 0.5567 & 0.5547 \\
\cline{3-15}  &   & \texttt{bm} & 0.5512 & 0.5184 & 0.5545 & 0.5618 & 0.5397 & 0.5592 & 0.5601 & 0.5940 & 0.5486 & 0.5531 & 0.5628 & 0.5486 \\
\cline{3-15}  &   & \texttt{jp} & 0.5719 & 0.5339 & 0.5647 & 0.5691 & 0.5337 & 0.5525 & 0.5701 & 0.6041 & 0.5567 & 0.5594 & 0.5729 & 0.5567 \\
\cline{3-15}  &   & \texttt{zh} & 0.5617 & 0.5240 & 0.5608 & 0.5616 & 0.5380 & 0.5529 & 0.5704 & 0.6026 & 0.5587 & 0.5622 & 0.5729 & 0.5587 \\
\hline 
\hline
\multicolumn{1}{|c|}{\multirow{10}{*}{\makecell{Weighted\\Regular\\BCE}}} & \multicolumn{1}{c|}{\multirow{5}{*}{\makecell{TOGA\\0.5563}}} & \texttt{en} & 0.6130 & 0.5682 & 0.5140 & 0.5219 & 0.5704 & 0.6329 & 0.5508 & 0.6054 & 0.5061 & 0.5041 & 0.5509 & 0.5665 \\
\cline{3-15}  &   & \texttt{fr} & 0.6027 & 0.5600 & 0.5589 & 0.6213 & 0.5404 & 0.6018 & 0.5574 & 0.6420 & 0.5101 & 0.4916 & 0.5539 & 0.5833 \\
\cline{3-15}  &   & \texttt{bm} & 0.6136 & 0.5915 & 0.5558 & 0.5907 & 0.5436 & 0.6149 & 0.5360 & 0.5698 & 0.5243 & 0.5455 & 0.5546 & 0.5825 \\
\cline{3-15}  &   & \texttt{jp} & 0.5902 & 0.5176 & 0.5627 & 0.6299 & 0.5569 & 0.6132 & 0.5636 & 0.6465 & 0.5243 & 0.4968 & 0.5595 & 0.5808 \\
\cline{3-15}  &   & \texttt{zh} & 0.6141 & 0.5844 & 0.5481 & 0.5687 & 0.5703 & 0.6114 & 0.5549 & 0.6090 & 0.5243 & 0.5437 & 0.5623 & 0.5834 \\
\cline{2-15}  & \multicolumn{1}{c|}{\multirow{5}{*}{\makecell{PIGA\\0.5759}}} & \texttt{en} & 0.6047 & 0.5632 & 0.5427 & 0.5388 & 0.5832 & 0.6139 & 0.5431 & 0.5698 & 0.5385 & 0.6069 & 0.5624 & 0.5785 \\
\cline{3-15}  &   & \texttt{fr} & 0.6303 & 0.6017 & 0.5455 & 0.5077 & 0.6045 & 0.6214 & 0.5624 & 0.5814 & 0.5547 & 0.6043 & 0.5795 & 0.5833 \\
\cline{3-15}  &   & \texttt{bm} & 0.6241 & 0.5961 & 0.5826 & 0.5219 & 0.5763 & 0.5926 & 0.5375 & 0.5306 & 0.5607 & 0.6226 & 0.5762 & 0.5728 \\
\cline{3-15}  &   & \texttt{jp} & 0.6313 & 0.5874 & 0.5619 & 0.5940 & 0.5941 & 0.6388 & 0.5550 & 0.6312 & 0.5486 & 0.5784 & 0.5782 & 0.6060 \\
\cline{3-15}  &   & \texttt{zh} & 0.6318 & 0.6360 & 0.5909 & 0.6008 & 0.5869 & 0.6220 & 0.5656 & 0.6379 & 0.5405 & 0.5540 & 0.5831 & 0.6101 \\
\hline
\hline
\multicolumn{2}{|c|}{\textit{\textbf{Trained Teacher Model}}} & \textit{\textbf{\texttt{en}}} & \textit{\textbf{0.7683}} & \textit{\textbf{0.7672}} & \textit{\textbf{0.7267}} & \textit{\textbf{0.7295}} & \textit{\textbf{0.6992}} & \textit{\textbf{0.7176}} & \textit{\textbf{0.6953}} & \textit{\textbf{0.7329}} & \textit{\textbf{0.6964}} & \textit{\textbf{0.6862}} & \textit{\textbf{0.7172}} & \textit{\textbf{0.7267}} \\
\hline
\multicolumn{1}{|c|}{\multirow{10}{*}{\makecell{Proposed\\CLAD}}} & \multicolumn{1}{c|}{\multirow{5}{*}{\makecell{TOGA\\0.6204}}} & \texttt{en} & 0.6559 & 0.6109 & 0.6642 & 0.6612 & 0.6120 & 0.6583 & 0.6024 & 0.5828 & 0.6174 & 0.5809 & 0.6304 & 0.6188 \\
\cline{3-15}  &   & \texttt{fr} & 0.6604 & 0.6101 & 0.6517 & 0.6574 & 0.6044 & 0.6667 & 0.6075 & 0.5950 & 0.6174 & 0.5695 & 0.6283 & 0.6197\\
\cline{3-15}  &   & \texttt{bm} & 0.6724 & 0.6525 & 0.6377 & 0.6398 & 0.5903 & 0.6325 & 0.5804 & 0.5570 & 0.6134 & 0.5527 & 0.6188 & 0.6069\\
\cline{3-15}  &   & \texttt{jp} & 0.6489 & 0.6171 & 0.6315 & 0.6360 & 0.5912 & 0.6454 & 0.5935 & 0.5641 & 0.6073 & 0.5650 & 0.6145 & 0.6055\\
\cline{3-15}  &   & \texttt{zh} & 0.6596 & 0.5899 & 0.6232 & 0.6353 & 0.6049 & 0.6620 & 0.5961 & 0.5774 & 0.5668 & 0.5286 & 0.6101 & 0.5987\\
\cline{2-15}  & \multicolumn{1}{c|}{\multirow{5}{*}{\makecell{PIGA\\0.6332}}} & \texttt{en} & 0.6619 & 0.6456 & 0.6439 & 0.6408 & 0.5819 & 0.6038 & 0.6174 & 0.6771 & 0.5951 & 0.5798 & 0.6201 & 0.6294\\
\cline{3-15}  &   & \texttt{fr} & 0.6729 & 0.6464 & 0.6537 & 0.6614 & 0.6111 & 0.6456 & 0.5989 & 0.6723 & 0.6012 & 0.5763 & 0.6276 & 0.6404\\
\cline{3-15}  &   & \texttt{bm} & 0.6749 & 0.6494 & 0.6638 & 0.6706 & 0.6152 & 0.6480 & 0.6296 & 0.6973 & 0.6215 & 0.5961 & 0.6410 & 0.6523\\
\cline{3-15}  &   & \texttt{jp} & 0.6746 & 0.6524 & 0.6539 & 0.6545 & 0.5946 & 0.6094 & 0.6131 & 0.6748 & 0.6235 & 0.6250 & 0.6319 & 0.6432\\
\cline{3-15}  &   & \texttt{zh} & 0.6819 & 0.6695 & 0.6660 & 0.6693 & 0.6020 & 0.6245 & 0.6166 & 0.6805 & 0.6356 & 0.6170 & 0.6404 & 0.6522\\
\hline
\end{tabular}

  \label{tab:results-essays}
  
\end{table*}

\begin{table*}
    \centering
    \scriptsize

    \caption{Performance comparison of the proposed CLAD algorithm with the zero-shot cross-lingual approach and the weighted BCE baseline across five languages (\texttt{en}, \texttt{fr}, \texttt{bm}, \texttt{jp}, \texttt{zh}), reporting Balanced Accuracy (BA) and F1-score for each personality dimension, evaluated on the Kaggle dataset.}
\begin{tabular}{|c|c|l|r|r|r|r|r|r|r|r|r|r|}
\hline
\multirow{2}{*}{\centering\textbf{Algorithm}} & \multirow{2}{*}{\centering\textbf{Dataset}} & \multicolumn{1}{c|}{\multirow{2}{*}{\centering\textbf{Language}}} & \multicolumn{2}{c|}{\textbf{O}} & \multicolumn{2}{c|}{\textbf{C}} & \multicolumn{2}{c|}{\textbf{E}} & \multicolumn{2}{c|}{\textbf{A}} & \multicolumn{2}{c|}{\textbf{Average}} \\
\cline{4-13}  &   &   & \multicolumn{1}{l|}{BA} & \multicolumn{1}{l|}{F1} & \multicolumn{1}{l|}{BA} & \multicolumn{1}{l|}{F1} & \multicolumn{1}{l|}{BA} & \multicolumn{1}{l|}{F1} & \multicolumn{1}{l|}{BA} & \multicolumn{1}{l|}{F1} & \multicolumn{1}{l|}{BA} & \multicolumn{1}{l|}{F1} \\
\hline\hline
\multicolumn{2}{|c|}{\textit{\textbf{Trained Multilingual Model	}}} & \textit{\textbf{\texttt{en}}} & \textit{\textbf{0.7294}} & \textit{\textbf{0.8932}} & \textit{\textbf{0.6997}} & \textit{\textbf{0.6159}} & \textit{\textbf{0.7243}} & \textit{\textbf{0.6073}} & \textit{\textbf{0.7625}} & \textit{\textbf{0.7846}} & \textit{\textbf{0.7290}} & \textit{\textbf{0.7253}}\\
\hline
\multirow{8}{*}{\makecell{Zero-\\Shot\\Cross-\\Lingual}} & \multirow{4}{*}{\makecell{TOGA\\0.5768}} & \texttt{fr} & 0.5855 & 0.8782 & 0.5830 & 0.4703 & 0.5491 & 0.2790 & 0.5841 & 0.6378 & \textit{0.5754} & \textit{0.5663} \\
\cline{3-13}  &   & \texttt{bm} & 0.5846 & 0.8814 & 0.5787 & 0.4617 & 0.5537 & 0.2853 & 0.5817 & 0.6357 & \textit{0.5747} & \textit{0.5660} \\
\cline{3-13}  &   & \texttt{jp} & 0.5947 & 0.8825 & 0.5840 & 0.4724 & 0.5492 & 0.2768 & 0.5797 & 0.6325 & \textit{0.5769} & \textit{0.5661} \\
\cline{3-13}  &   & \texttt{zh} & 0.6096 & 0.8869 & 0.5813 & 0.4657 & 0.5489 & 0.2799 & 0.5805 & 0.6321 & \textit{0.5801} & \textit{0.5662} \\
\cline{2-13}  & \multirow{4}{*}{\makecell{PIGA\\0.6171}} & \texttt{fr} & 0.6229 & 0.8937 & 0.6155 & 0.5108 & 0.6012 & 0.3761 & 0.6199 & 0.6697 & \textit{0.6149} & \textit{0.6126} \\
\cline{3-13}  &   & \texttt{bm} & 0.6215 & 0.8942 & 0.6160 & 0.5123 & 0.6016 & 0.3766 & 0.6247 & 0.6708 & \textit{0.6160} & \textit{0.6135} \\
\cline{3-13}  &   & \texttt{jp} & 0.6235 & 0.8964 & 0.6073 & 0.5004 & 0.6009 & 0.3739 & 0.6387 & 0.6828 & \textit{0.6176} & \textit{0.6134} \\
\cline{3-13}  &   & \texttt{zh} & 0.6363 & 0.8986 & 0.6122 & 0.5091 & 0.6044 & 0.3811 & 0.6272 & 0.6772 & \textit{0.6200} & \textit{0.6165} \\
\hline
\hline
\multicolumn{1}{|c|}{\multirow{10}{*}{\makecell{Weighted\\Regular\\BCE}}} & \multicolumn{1}{c|}{\multirow{5}{*}{\makecell{TOGA\\0.6516}}} & \texttt{en} & 0.6527 & 0.8631 & 0.6078 & 0.5394 & 0.5925 & 0.3668 & 0.7261 & 0.7529 & 0.6448 & 0.6305 \\
\cline{3-13}  &   & \texttt{fr} & 0.6591 & 0.8598 & 0.6032 & 0.5336 & 0.6187 & 0.4123 & 0.7281 & 0.7473 & 0.6523 & 0.6382 \\
\cline{3-13}  &   & \texttt{bm} & 0.6790 & 0.8595 & 0.6024 & 0.5288 & 0.6036 & 0.3882 & 0.7216 & 0.7526 & 0.6516 & 0.6323 \\
\cline{3-13}  &   & \texttt{jp} & 0.6800 & 0.8607 & 0.6115 & 0.5462 & 0.6068 & 0.3922 & 0.7192 & 0.7527 & 0.6544 & 0.6379 \\
\cline{3-13}  &   & \texttt{zh} & 0.6561 & 0.8561 & 0.6207 & 0.5576 & 0.6201 & 0.4159 & 0.7237 & 0.7445 & 0.6552 & 0.6435 \\
\cline{2-13}  & \multicolumn{1}{c|}{\multirow{5}{*}{\makecell{PIGA\\0.6780}}} & \texttt{en} & 0.6744 & 0.8627 & 0.6263 & 0.5580 & 0.6193 & 0.4112 & 0.7506 & 0.7687 & 0.6676 & 0.6502 \\
\cline{3-13}  &   & \texttt{fr} & 0.6985 & 0.8698 & 0.6492 & 0.5869 & 0.6375 & 0.4415 & 0.7617 & 0.7726 & 0.6867 & 0.6677 \\
\cline{3-13}  &   & \texttt{bm} & 0.6792 & 0.8641 & 0.6208 & 0.5468 & 0.6275 & 0.4256 & 0.7353 & 0.7509 & 0.6657 & 0.6469 \\
\cline{3-13}  &   & \texttt{jp} & 0.7042 & 0.8678 & 0.6411 & 0.5777 & 0.6449 & 0.4536 & 0.7546 & 0.7600 & 0.6862 & 0.6648 \\
\cline{3-13}  &   & \texttt{zh} & 0.6929 & 0.8717 & 0.6388 & 0.5740 & 0.6472 & 0.4574 & 0.7563 & 0.7575 & 0.6838 & 0.6652 \\
\hline
\hline
\multicolumn{2}{|c|}{\textit{\textbf{Trained Teacher Model}}} & \textit{\textbf{\texttt{en}}} & \textit{\textbf{0.8702}} & \textit{\textbf{0.7648}} & \textit{\textbf{0.7267}} & \textit{\textbf{0.7295}} & \textit{\textbf{0.6992}} & \textit{\textbf{0.7176}} & \textit{\textbf{0.6953}} & \textit{\textbf{0.7329}} & \textit{\textbf{0.7479}} & \textit{\textbf{0.7362}} \\
\hline
\multicolumn{1}{|c|}{\multirow{10}{*}{\makecell{Proposed\\CLAD}}} & \multicolumn{1}{c|}{\multirow{5}{*}{\makecell{TOGA\\0.7305}}} & \texttt{en} & 0.7250 & 0.8197 & 0.6950 & 0.6324 & 0.7113 & 0.5648 & 0.7805 & 0.8078 & 0.7279 & 0.7062 \\
\cline{3-13}  &   & \texttt{fr} & 0.7169 & 0.8137 & 0.6772 & 0.6070 & 0.7115 & 0.5641 & 0.7858 & 0.8055 & 0.7228 & 0.6976 \\
\cline{3-13}  &   & \texttt{bm} & 0.7306 & 0.8197 & 0.6780 & 0.6087 & 0.7200 & 0.5764 & 0.7829 & 0.8046 & 0.7279 & 0.7023 \\
\cline{3-13}  &   & \texttt{jp} & 0.7471 & 0.8292 & 0.6847 & 0.6157 & 0.7307 & 0.5929 & 0.7816 & 0.8021 & 0.7360 & 0.7100 \\
\cline{3-13}  &   & \texttt{zh} & 0.7371 & 0.8258 & 0.7007 & 0.6374 & 0.7231 & 0.5842 & 0.7898 & 0.8109 & 0.7377 & 0.7146 \\
\cline{2-13}  & \multicolumn{1}{c|}{\multirow{5}{*}{\makecell{PIGA\\0.7448}}} & \texttt{en} & 0.7336 & 0.8237 & 0.7089 & 0.6586 & 0.7356 & 0.6008 & 0.7941 & 0.8209 & 0.7430 & 0.7260 \\
\cline{3-13}  &   & \texttt{fr} & 0.7589 & 0.8399 & 0.7068 & 0.6568 & 0.7408 & 0.6095 & 0.7869 & 0.8136 & 0.7483 & 0.7300 \\
\cline{3-13}  &   & \texttt{bm} & 0.7235 & 0.8249 & 0.6916 & 0.6397 & 0.7324 & 0.5928 & 0.7929 & 0.8166 & 0.7351 & 0.7185 \\
\cline{3-13}  &   & \texttt{jp} & 0.7539 & 0.8334 & 0.6879 & 0.6340 & 0.7508 & 0.6240 & 0.7866 & 0.8142 & 0.7448 & 0.7264 \\
\cline{3-13}  &   & \texttt{zh} & 0.7638 & 0.8391 & 0.7104 & 0.6608 & 0.7428 & 0.6156 & 0.7937 & 0.8184 & 0.7527 & 0.7335 \\
\hline

\end{tabular}%
 
    \label{tab:results-kaggle}
    
  \end{table*}

\subsection{PIGA Analysis}

We evaluate PIGA by comparing it with human-written text from the original dataset and with TOGA, which we identify as the fairest baseline relative to standard machine translation systems. Conventional translation services often fail to preserve deeper semantic intent, particularly in instances involving symbolism, rhetorical questions, or implicit cues, leading to information loss and thus unfair comparisons. To ensure methodological rigor, we conduct a comprehensive linguistic analysis examining stylistic alignment, semantic fidelity, and personality expression, alongside a systematic performance evaluation to assess downstream task implications. Although large language models may partially capture personality nuances and thus appear comparable to PIGA, even subtle divergences reveal the added value of PIGA beyond simple dataset memorization or pattern recollection. Finally, we explicitly analyze potential data leakage risks to ensure that observed performance gains stem from genuine modeling improvements rather than inadvertent overlap with training data.

\subsubsection{Linguistic Study}
\begin{figure*}
    \centering
    \subfigure[Syllable Count in Essays Dataset]{
        \includegraphics[width=0.31\textwidth]{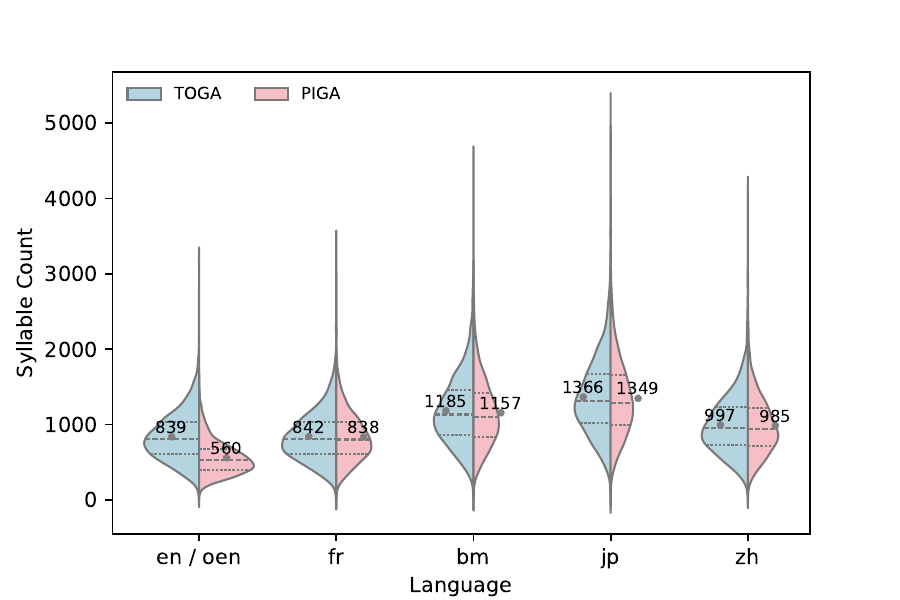}
        \label{fig:sc-essays}
    }
    \subfigure[Herdan's C index in Essays Dataset]{
        \includegraphics[width=0.31\textwidth]{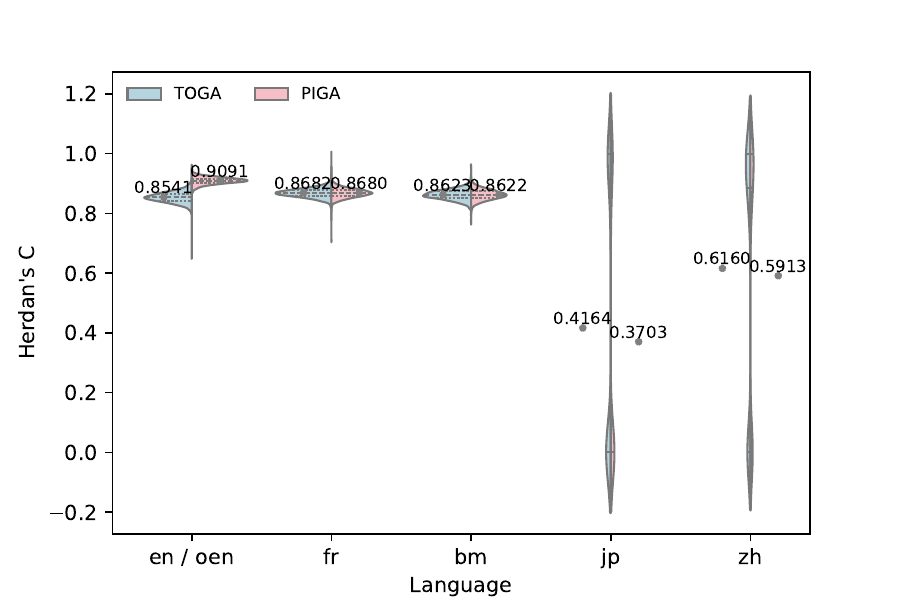}
        \label{fig:hc-essays}
    }
    \subfigure[Lexical Density in Essays Dataset]{
        \includegraphics[width=0.31\textwidth]{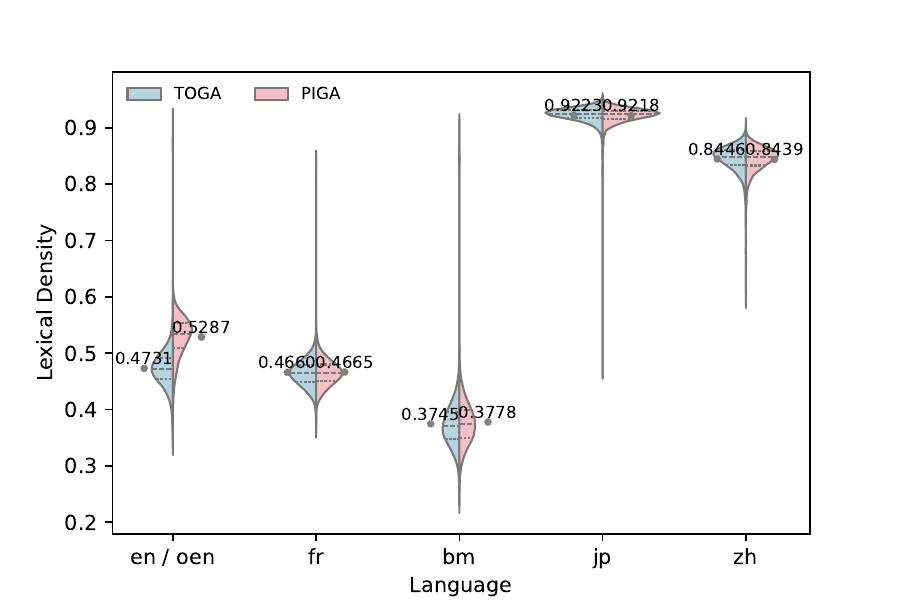}
        \label{fig:ld-essays}
    }
    \subfigure[Syllable Count in Kaggle Dataset]{
        \includegraphics[width=0.31\textwidth]{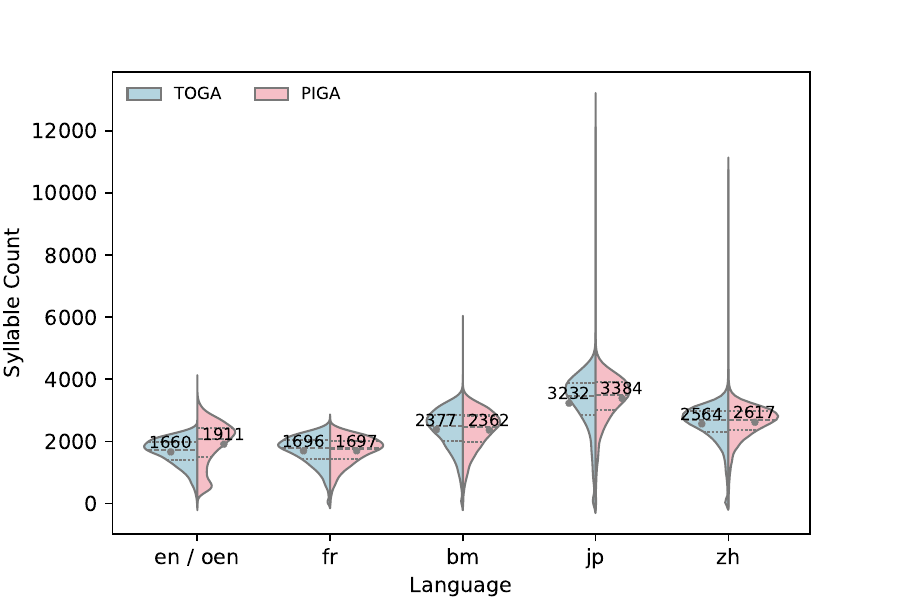}
        \label{fig:sc-kaggle}
    }
    \subfigure[Herdan's C index in Kaggle Dataset]{
        \includegraphics[width=0.31\textwidth]{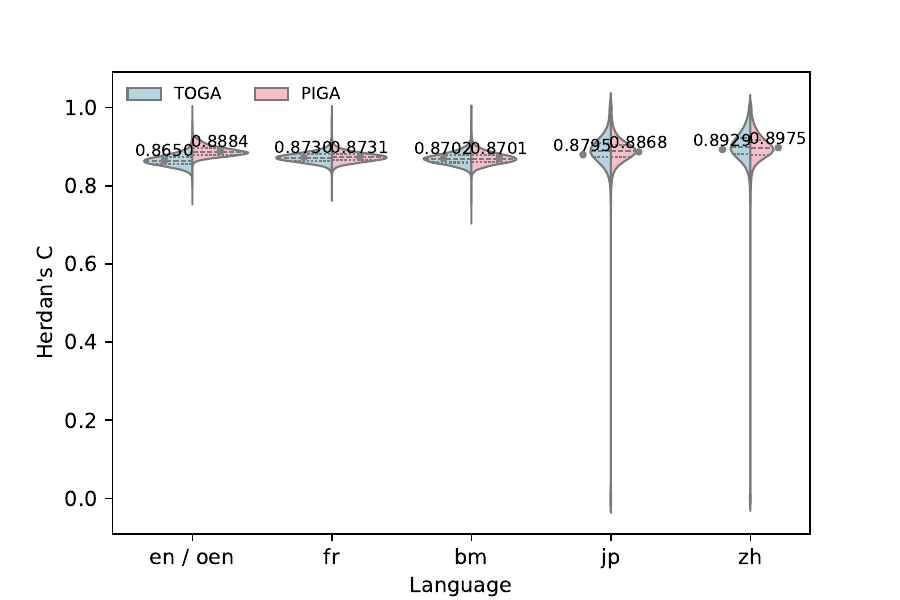}
        \label{fig:hc-kaggle}
    }
    \subfigure[Lexical Density in Kaggle Dataset]{
        \includegraphics[width=0.31\textwidth]{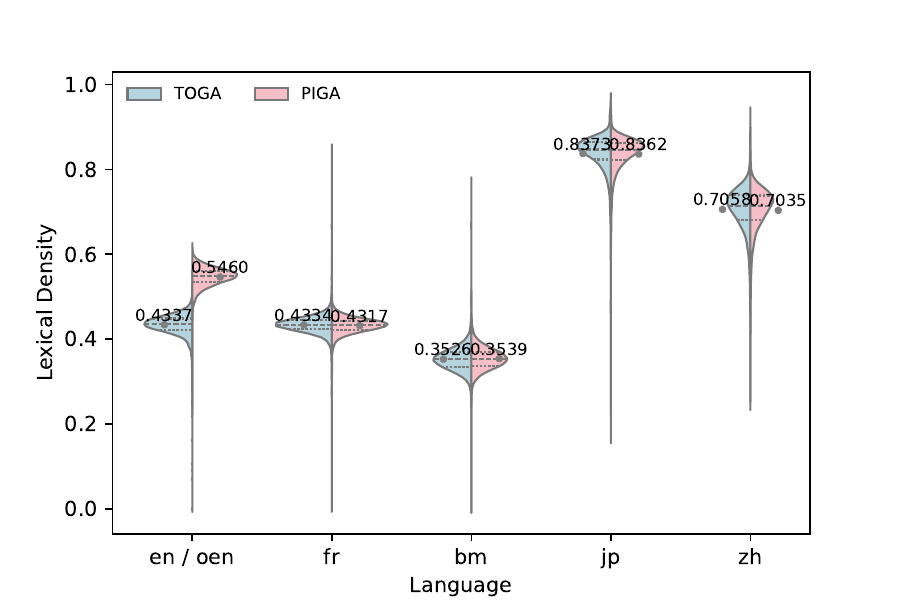}
        \label{fig:ld-kaggle}
    }
    \caption{Comparison of TOGA and PIGA Translation Approaches in Essays and Kaggle Datasets Across Five Languages (\texttt{en}, \texttt{fr}, \texttt{bm}, \texttt{jp}, \texttt{zh}) using the \texttt{gpt-4o-2024-08-06} model. Figures \ref{fig:sc-essays} and \ref{fig:sc-kaggle} show the syllable count distributions, while Figures \ref{fig:hc-essays} and \ref{fig:hc-kaggle} present Herdan's C index. Figures \ref{fig:ld-essays} and \ref{fig:ld-kaggle} illustrate lexical density for the respective datasets. However, no translation is required for the English dataset. To provide a more intuitive comparison—where one version is original and the other is ``translated"—we included a fully complementary version of the original English dataset text, annotated with labels reflecting terminology, grammar, and style. For TOGA, \texttt{en} represents the original dataset, while for PIGA, we present the opposite personality augmentation (\texttt{oen}). Since this aspect requires prior content analysis, we adopted OpenAI's reasoning LLM, \texttt{o3-mini-2025-01-31}. In general, the trend remains consistent across both datasets, regardless of the metric used.
}
    \label{fig:piga}
\end{figure*}

We visualized the Syllable Count (SC), Herdan’s C index (HC), and Lexical Density (LD) in Fig. \ref{fig:piga}. SC is commonly used to estimate text complexity and phonetic richness \cite{Spitzley2022,Mehta2020}. HC measures vocabulary diversity, reflecting lexical richness in a given text \cite{Herdan2006,Shah2023}. LD indicates the proportion of content words, which helps assess linguistic compactness and informativeness \cite{Zheng2025}. We expect PIGA to maintain consistency across these metrics while enhancing personality retention in translated texts in both datasets.

In the Essays dataset, SC decreases with PIGA, whereas in the Kaggle dataset, it increases. This discrepancy stems from the inherent imbalance in the Kaggle dataset, where certain personality dimensions have a stronger influence on SC. For example, the Openness trait, associated with exploring ideas, tends to generate more syllables. Since over 80\% of the Kaggle dataset falls under this category, it skews the SC trend. However, research suggests that contextual outcomes are not solely determined by a single personality dimension, making the Kaggle dataset less coherent due to its imbalance \cite{Mehta2020}.

On the other hand, we observe a similar pattern in both datasets in the distribution of HC and LD scores. This demonstrates that PIGA remains consistent in augmentation and does not significantly alter contextual meaning. Since HC measures vocabulary diversity and LD reflects content compactness, their stability across datasets suggests that PIGA successfully retains key stylistic and lexical traits while translating. The observed differences may also be influenced by linguistic characteristics. For example, the LD scores for \texttt{fr} and \texttt{bm} are slightly higher in PIGA, likely because word choice plays a crucial role in expressing personality in these languages. On the other hand, for \texttt{jp} and \texttt{zh}, personality reflection is more dependent on sentence structure rather than individual word selection. This suggests that PIGA adapts to linguistic characteristics while maintaining consistency in augmentation.

Moreover, in French (\texttt{fr}), personality-driven expressions often involve richer adjectives or verb variations, while in Japanese (\texttt{jp}), shifts in formality levels and syntactic structures may play a bigger role. These findings highlight that PIGA not only preserves personality traits consistently but also adapts to the linguistic nuances of different languages. While this demonstrates the robustness of PIGA, future research could further explore its handling of idiomatic expressions and cultural nuances to refine its augmentation effectiveness.

In order to showcase the effect of the PIGA method, we focus on the English language where we make an opposite personality augmentation (\texttt{oen}) on the original english content (\texttt{en}, shown in TOGA categories). We observed a similar trend in HC and LD as well, reflecting the original dataset (\texttt{en}), which was already personalized (collected from real humans). The \texttt{oen} further demonstrates that the LLM managed to produce a personalized text, as it exhibits a more intuitive difference. As aforementioned, the Kaggle dataset with imbalanced SC values does not show a similar trend (the Essays dataset increases, while the Kaggle dataset decreases). This phenomenon becomes more evident in \texttt{en} and \texttt{oen}, further justifying that the PIGA augmentation generates better augmentation for multilingual personality recognition.

\subsubsection{Performance Implications}
As shown in Tables~\ref{tab:results-essays} and~\ref{tab:results-kaggle}, PIGA demonstrates consistently superior performance compared to TOGA across both the Essays and Kaggle datasets, regardless of the underlying training algorithm (which will be detailed in the next subsection). In the zero-shot cross-lingual experiments, where the model is trained exclusively on English data and subsequently evaluated on other languages, the PIGA dataset provides a robust margin of improvement. Because these experiments are not influenced by multilingual training, the effect of augmentation becomes more apparent, yielding an improvement of approximately 3\% in Essays dataset and 4\% in Kaggle dataset. 

This trend persists when the model is trained on multiple languages using BCE loss, with gains of approximately 2\% on the Essays dataset and 2.6\% on Kaggle, as well as under the proposed CLAD approach, which achieves improvements of about 1.3\% on Essays and 1.4\% on Kaggle. With a $p$-value below 0.05 in Figure~\ref{fig:stats}, these results indicate a statistically significant advancement over those obtained with the TOGA dataset under the same experimental settings. 

The consistent performance of the PIGA dataset across all algorithms underscores its reliability as a superior translation resource that preserves richer psychological signals, thereby enhancing personality recognition accuracy.

\subsubsection{Label Leakage Study}
However, we acknowledge that PIGA has a potential \textbf{limitation}, where it may introduce unintended biases to some extent. Nevertheless, this is the best trade-off available, as no human is involved in generating multilingual augmentation. Due to concerns that the PIGA method may have exploited the labels during the augmentation process, we will include results from both methods' datasets in the evaluation for CLAD. To investigate this issue, we conducted a series of probing experiment (i) LLM as Probing Judges, and (ii) Probing Classifier on Embeddings.

\begin{figure*}
    \centering
    \subfigure[Content Checking Prompt]{
        \includegraphics[width=0.45\textwidth]{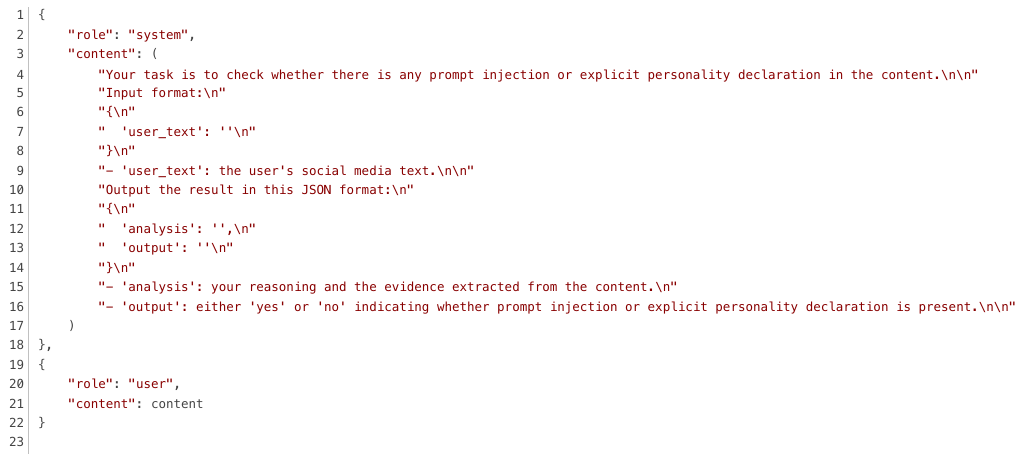}
        \label{fig:checkprompt}
    }
    \subfigure[Translation Comparison Prompt]{
        \includegraphics[width=0.45\textwidth]{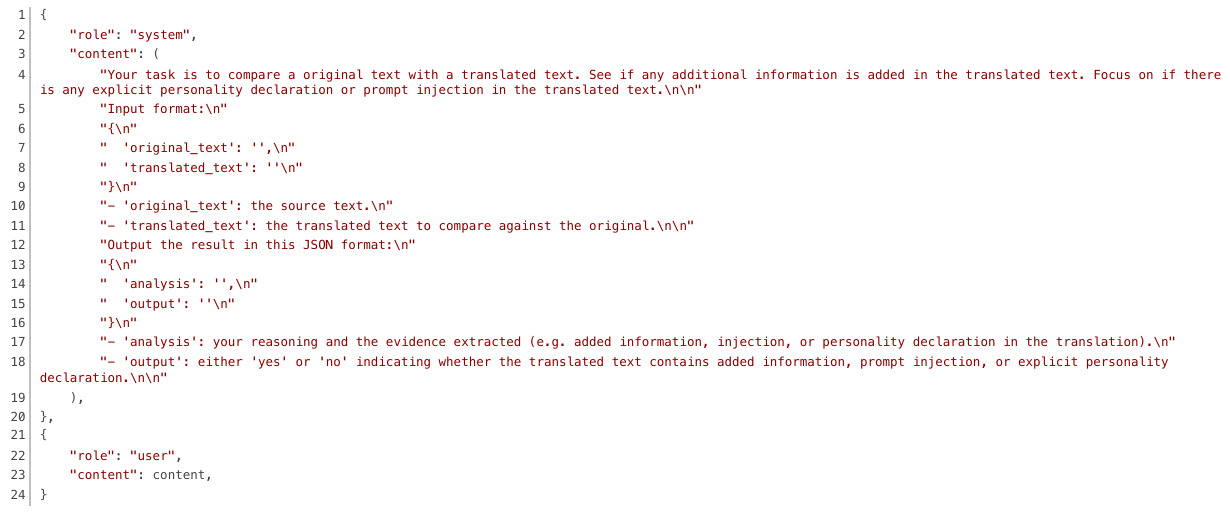}
        \label{fig:compareprompt}
    }
\caption{Prompts used to assess label leakage in the ``LLMs as Probing Judges" evaluation.
Fig. \ref{fig:checkprompt} is the Content Screening Prompt (CSP), which asks the LLM to determine whether the translated text explicitly expresses the personality label or contains direct trait-related cues. 
Fig. \ref{fig:compareprompt} is the Translation Comparison Prompt (TCP), which instructs the LLM to compare the original and translated texts to identify any additional trait-aligned information, semantic enrichment, or stylistic amplification introduced during augmentation.
}
    \label{fig:prompt}
\end{figure*}

\begin{figure*}
    \centering
    \subfigure[Essays Dataset (CSP)]{
        \includegraphics[width=0.23\textwidth]{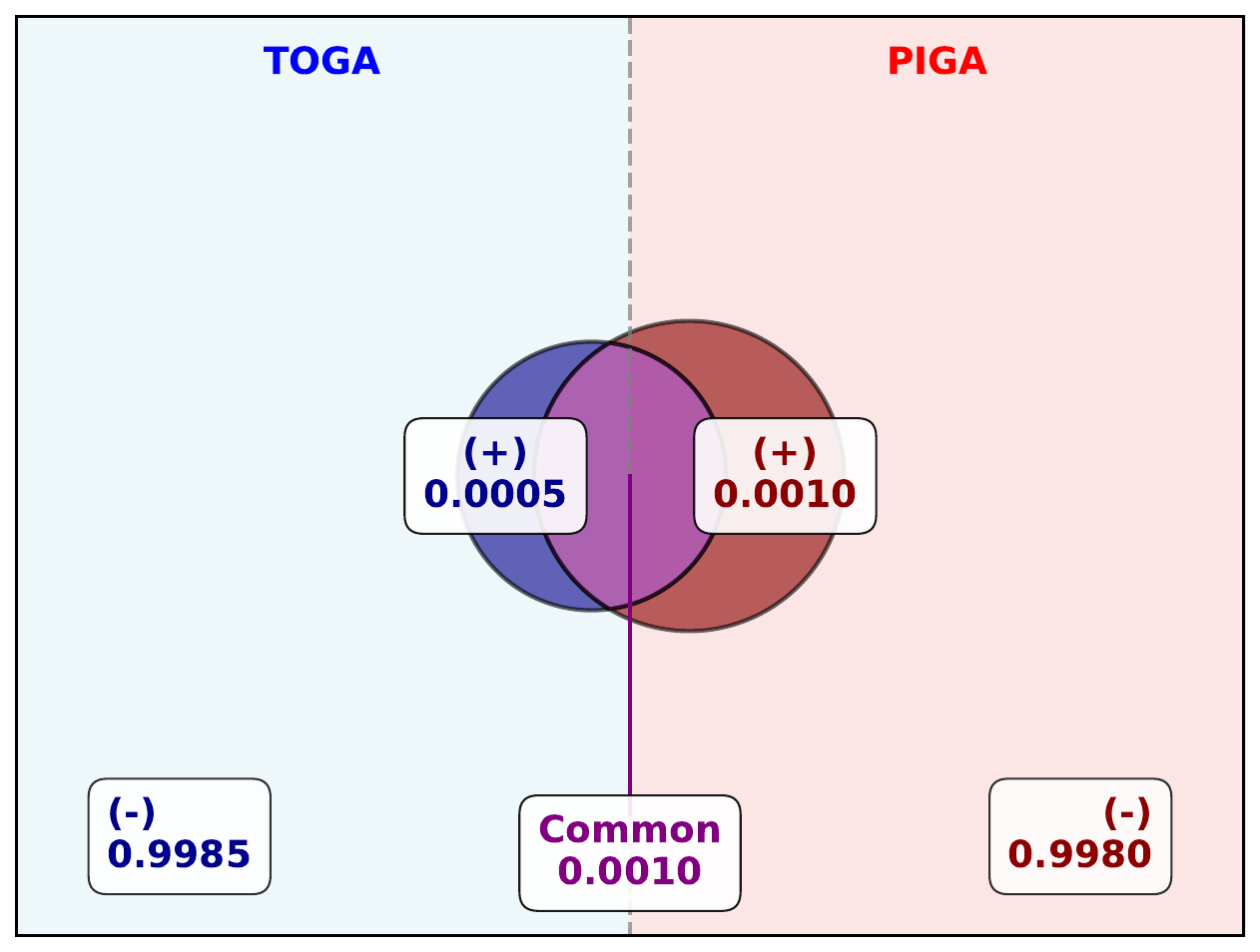}
        \label{fig:essays_venn}
    }
    \subfigure[Kaggle Dataset (CSP)]{
        \includegraphics[width=0.23\textwidth]{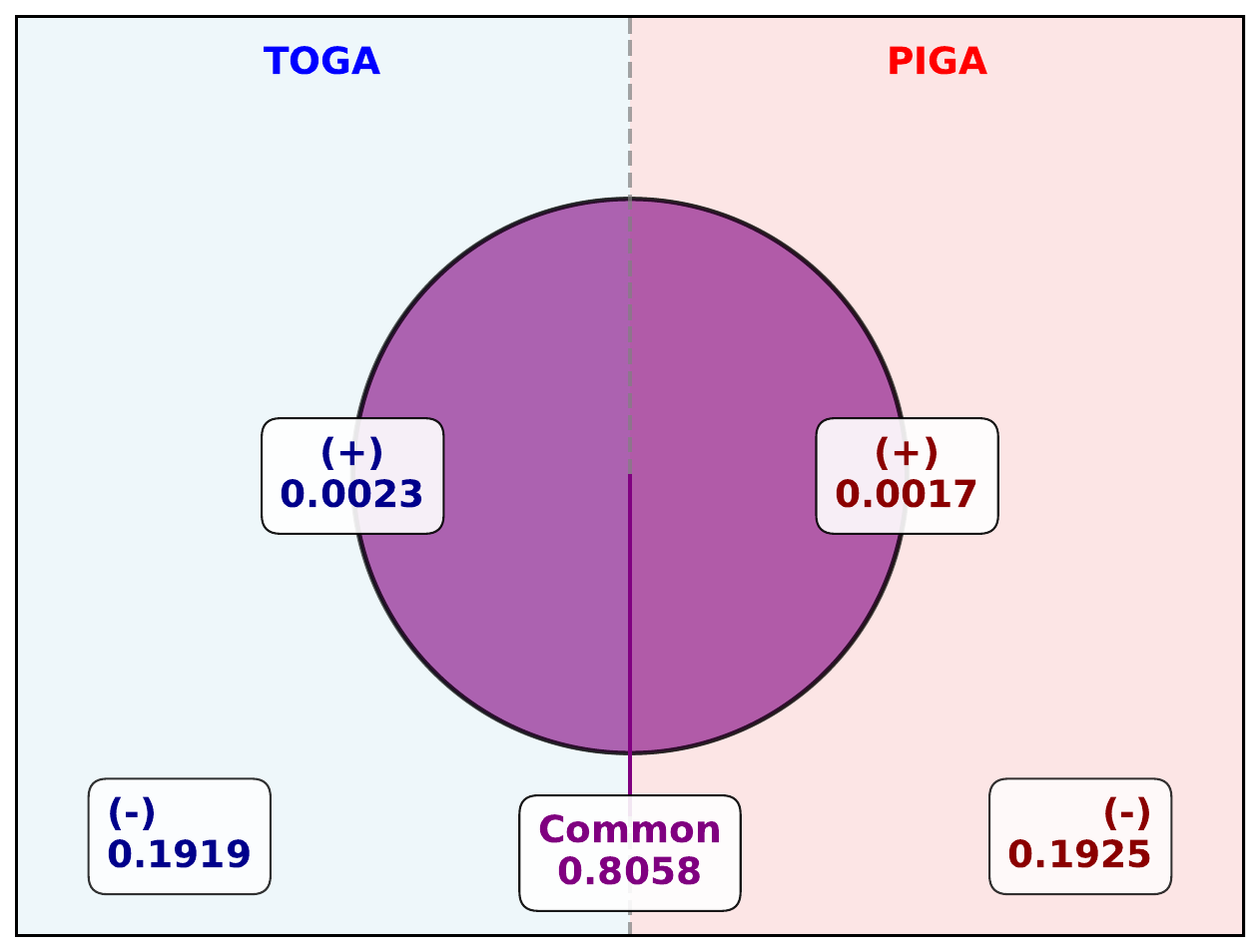}
        \label{fig:kaggle_venn}
    }
    \subfigure[Essays Dataset (TCP)]{
        \includegraphics[width=0.23\textwidth]{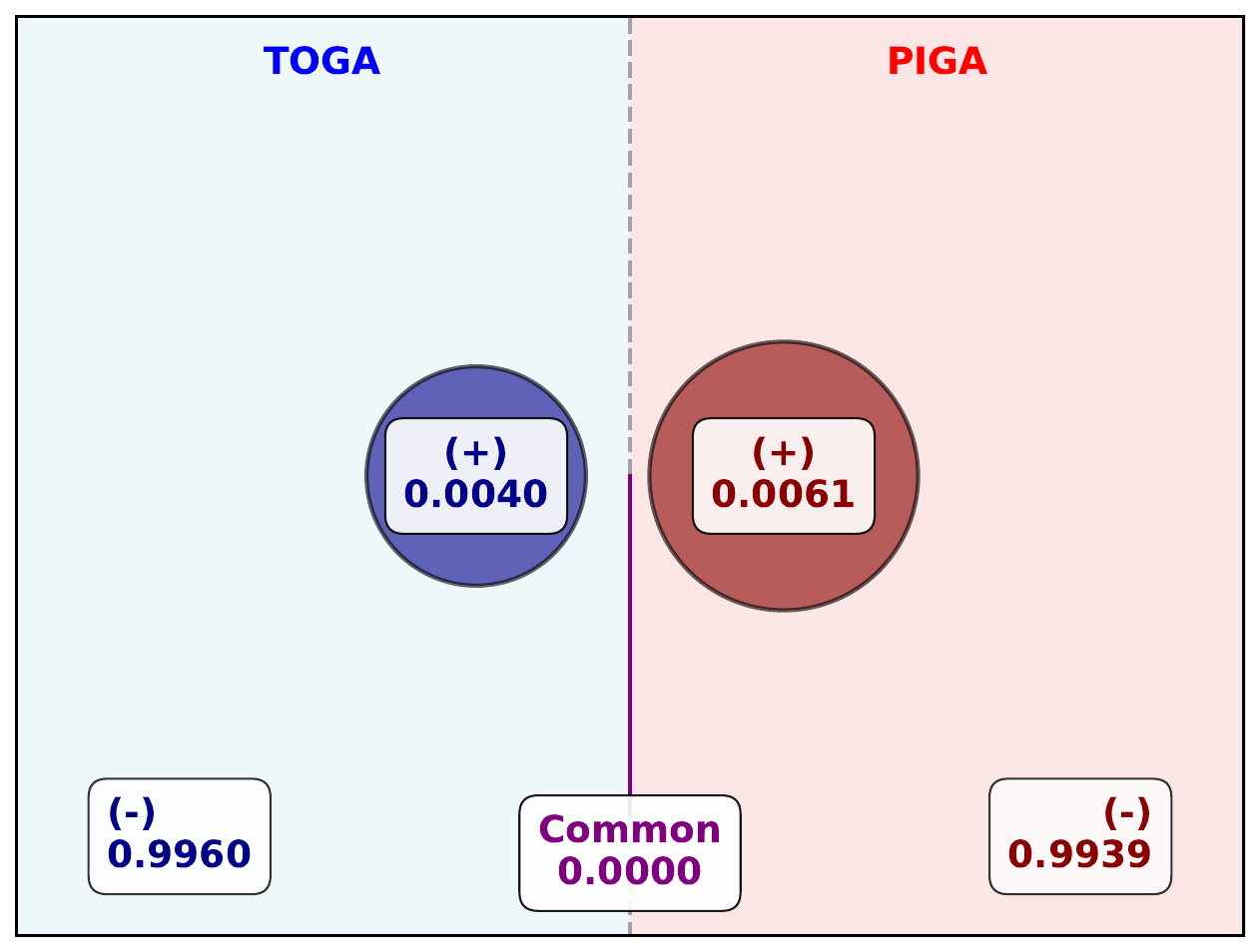}
        \label{fig:essays_vennadd}
    }
    \subfigure[Kaggle Dataset (TCP)]{
        \includegraphics[width=0.23\textwidth]{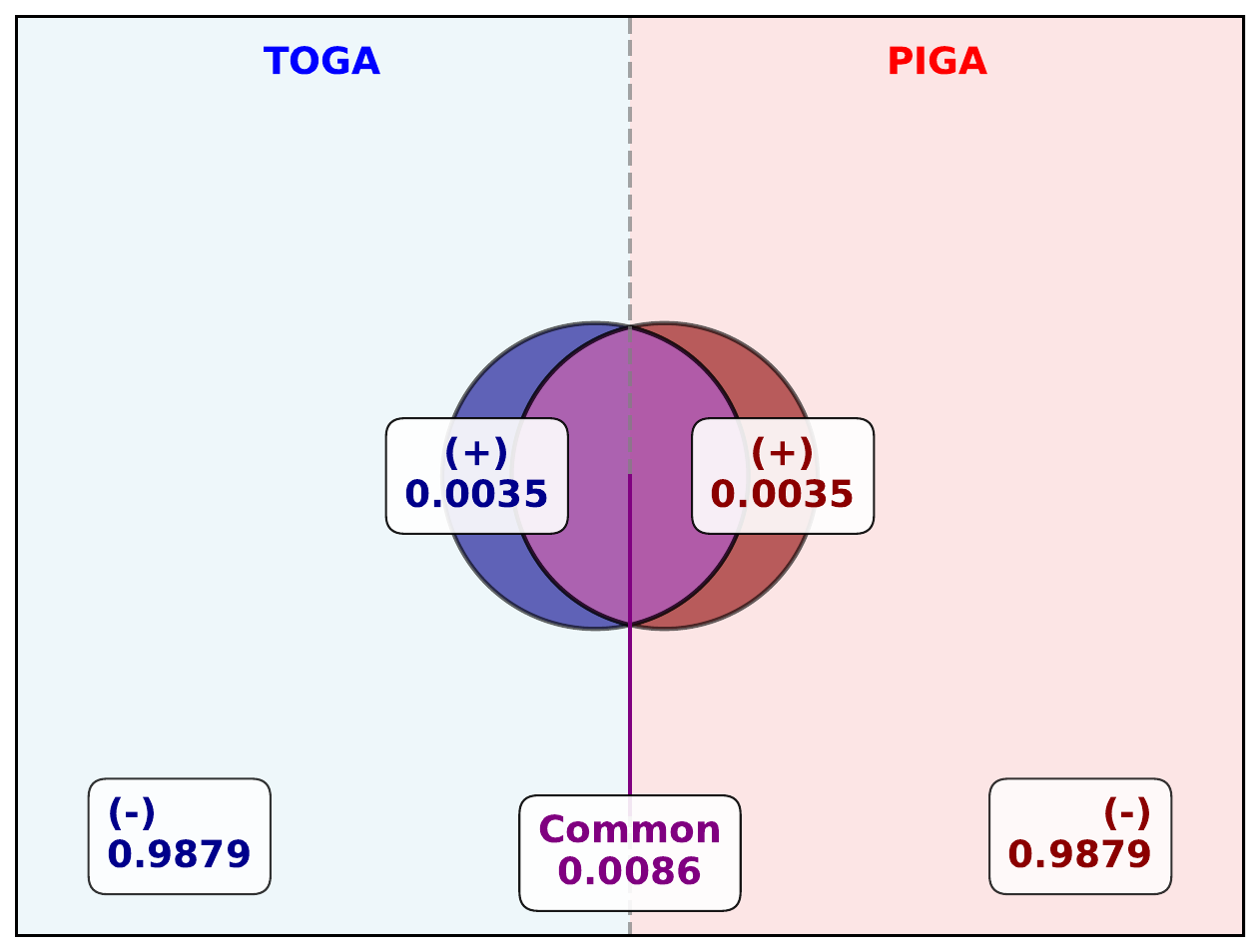}
        \label{fig:kaggle_vennadd}
    }
    
    \caption{Venn diagram illustrating the LLM's evaluation of whether explicit instructions induce label leakage on the Essays and Kaggle datasets using 2 designed prompts as shown in Fig \ref{fig:prompt}. Each circle represents instances flagged as exhibiting label leakage (+), while non-leakage cases are denoted by (-). The overlapping region (``Common'') indicates instances occurring in the same dataset rows in both TOGA and PIGA.}
    \label{fig:venn}
\end{figure*}

\textbf{LLMs as Probing Judges}. As shown in Figure~\ref{fig:prompt}, we design two prompts---the Translation Comparison Prompt (TCP) and the Content Screening Prompt (CSP)---to instruct the LLM to evaluate the translated text and determine whether it contains explicit expressions of the personality label or introduces additional trait-aligned information beyond the original content. The evaluation is visualized in Fig. \ref{fig:venn}. Since TOGA dataset is generated without access to personality labels, it serves as a control condition, providing a baseline for a reasonable or expected rate of flagged instances. We additionally report the intersection between TOGA and PIGA, representing identical rows flagged in both datasets. These overlapping cases likely reflect inherently ambiguous content that may prompt the LLM to raise false positives, rather than genuine label leakage introduced during augmentation.

With this approach, both the CSP and TCP consistently indicate that the Essays dataset is effectively free from label leakage. The proportion of flagged samples is close to zero under both evaluation prompts, suggesting that neither explicit trait expressions nor augmentation-induced trait-aligned information is present.

In contrast, the Kaggle dataset exhibits a relatively high proportion of flagged instances in both TOGA and PIGA. This pattern is expected, as the dataset is derived from personality-focused discussions and therefore naturally contains trait-related content. Importantly, this interpretation is supported by the TOGA results: the substantial overlap between TOGA and PIGA indicates that most flagged cases stem from pre-existing, label-relevant discourse rather than augmentation artifacts. Crucially, the potential concern of augmentation-induced leakage is mitigated by the very low proportion of uniquely flagged cases under the TSP, where the majority of flagged instances fall within the shared ``common" subset. This suggests that the flagged content primarily reflects inherent characteristics of the source material rather than systematic leakage introduced during translation or augmentation.

\begin{figure}
    \centering
    \subfigure[Essays Dataset]{
        \includegraphics[width=0.45\linewidth]{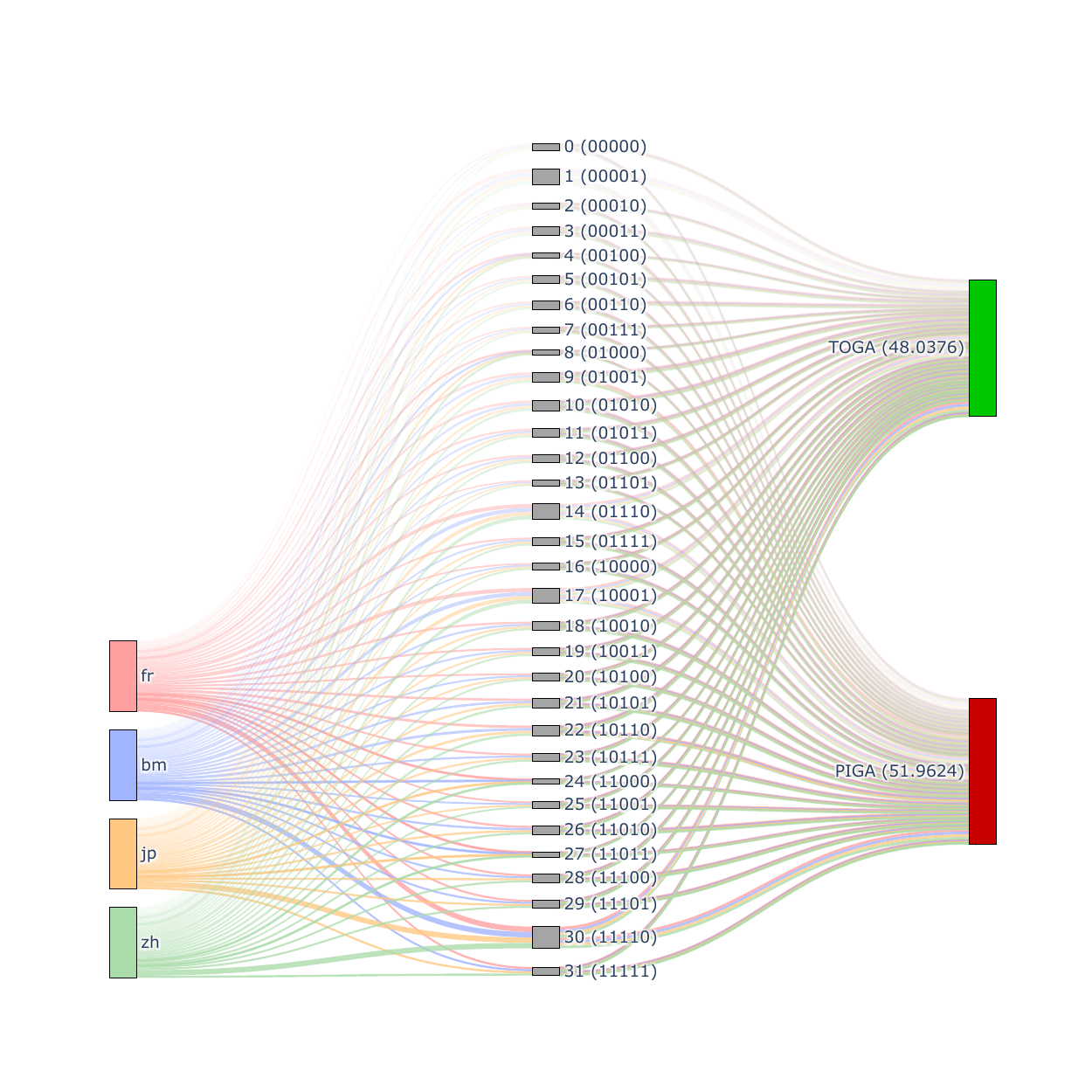}
        \label{fig:essays_probing}
    }
    \subfigure[Kaggle Dataset]{
        \includegraphics[width=0.45\linewidth]{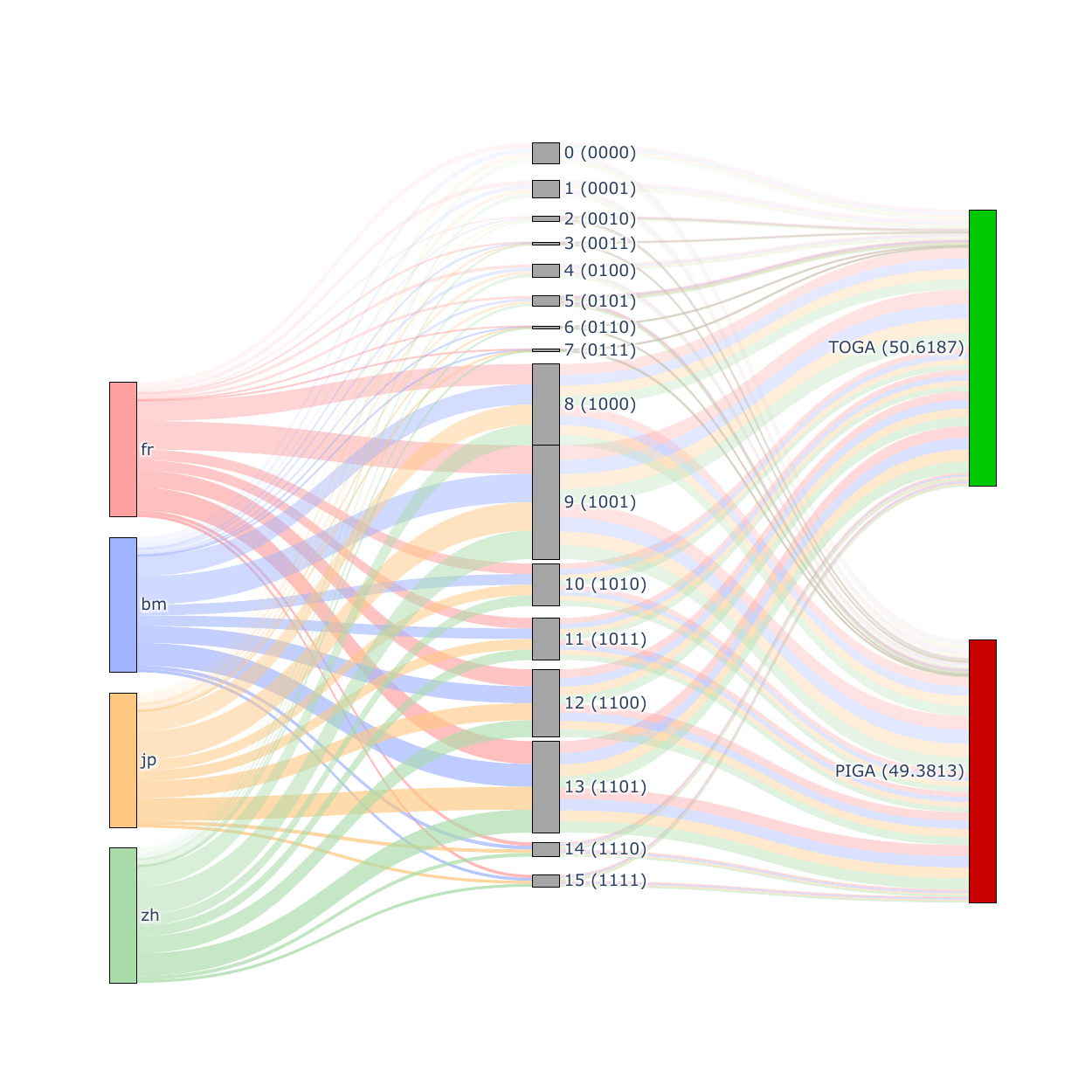}
        \label{fig:kaggle_probing}
    }
    \caption{
        Sankey diagrams illustrating the probing classifier results on CLAD embeddings for four augmented languages (\texttt{fr}, \texttt{bm}, \texttt{jp}, \texttt{zh}) using Essays Dataset (\ref{fig:essays_probing}) and Kaggle Dataset (\ref{fig:kaggle_probing}). The flow proceeds from language groups to personality types (middle layer), and finally to the binary classification outputs: TOGA and PIGA. The probing classifier is implemented using logistic regression with L2 regularization and trained with binary cross-entropy (BCE) loss. No clear separability trend is observed across both datasets, suggesting the absence of strong label-specific signals within the embeddings.
}
    \label{fig:probing}
\end{figure}

\textbf{Probing Classifier on Embeddings}, to examine whether the LLM induces systematic patterns in the embedding space that make personality labels more easily predictable—thereby potentially encouraging downstream models to rely on spurious or augmentation-induced signals rather than authentic personality cues—we conduct a $2 \times 2$ probing setup. Specifically, we evaluate two probing classifiers, Logistic Regression (LR) and Support Vector Machine (SVM), across two embedding sources: (i) the transformer encoder of our trained ADAM model, and (ii) the closed-source embedding model (\texttt{text-embedding-ada-002}) provided by OpenAI. Each probing model was trained to distinguish between PIGA and TOGA text embeddings, using the same train--validation--test split for consistency across all configurations. Since this is a binary classification task, the ideal outcome under the absence of systematic differences is performance close to \textbf{50\% accuracy}. Results substantially above this threshold would suggest that structured distinctions remain in the embedding space. To ensure robustness, we repeated each configuration across 5 independent runs with different random seeds. The aggregated results are reported in Table~\ref{tab:probing}, while Fig.~\ref{fig:probing} illustrates the distribution from one representative experiment.

\begin{table*}
\centering
\scriptsize
\caption{Probing classification results (mean, standard deviation, and 95\% confidence interval) on the Essays and Kaggle datasets using two probing models (Logistic Regression and SVM) over embeddings derived from the ADAM transformer encoder and OpenAI’s \textit{text-embedding-ada-002} model.}
\begin{tabular}{|c|c|c|c|c|l|l|}
\hline
\multicolumn{1}{|r|}{\multirow{2}{*}{\textbf{Datasets}}} &
\multirow{2}{*}{\textbf{Variation}} &
\multicolumn{2}{c|}{\textbf{Mean}} &
\multirow{2}{*}{\textbf{Standard Deviation}} &
\multicolumn{2}{c|}{\textbf{95\% Confidence Interval}} \\
\cline{3-4} \cline{6-7}
 &  & (-) & (+) &  & (-) & (+) \\
\hline
\multicolumn{1}{|c|}{\multirow{4}[1]{*}{Essays}} & LR (ADAM) & 49.4922 & 50.5078 & 0.9943 & [48.6207, 50.3637] & [49.6363, 51.3793] \\
\cline{2-7}  & LR (OpenAI) & 48.7646 & 51.2354 & 1.5403 & [47.4145, 50.1147] & [49.8853, 52.5855] \\
\cline{2-7}  & SVM (ADAM) & 49.8169 & 50.1831 & 1.8581 & [48.1883, 51.4456] & [48.5544, 51.8117]\\
\cline{2-7}  & SVM (OpenAI) & 51.2379 & 48.7621 & 1.8431 & [49.6223, 52.8534] & [47.1466, 50.3777] \\
\hline
\multicolumn{1}{|c|}{\multirow{4}[1]{*}{Kaggle}} & LR (ADAM) & 50.3461 & 49.6539 & 0.7904 & [49.6533, 51.0389] & [48.9611, 50.3467] \\
\cline{2-7}  & LR (OpenAI) & 49.9929 & 50.0071 & 0.7976 & [49.2937, 50.6920] & [49.3080, 50.7063] \\
\cline{2-7}  & SVM (ADAM) & 49.9466 & 50.0534 & 0.8035 & [49.2423, 50.6509] & [49.3491, 50.7577] \\
\cline{2-7}  & SVM (OpenAI) & 50.2022 & 49.7978 & 1.1560 & [49.1889, 51.2155] & [48.7845, 50.8111] \\
\hline
\end{tabular}
\label{tab:probing}
\end{table*}

From the results in Table \ref{tab:probing}, we observe that the classifier's performance remains close to the ideal 50\% accuracy threshold, regardless of which embedding model or recognition model is used. There is no noticeable trend across the probing models adopted or the embedding models employed. On the other hand, we notice that the Kaggle dataset is relatively closer to the 50\% threshold. This observation aligns with the nature of the Essays dataset, which was collected in a more constrained and structured environment that may influence the way personality traits are expressed. Such controlled settings could introduce subtle biases in how personality is reflected through the text, potentially affecting the embedding space. As a result, the probing classifier on the Essays dataset shows a slightly higher deviation from the ideal 50\% split, indicating a very weak but noticeable signal. Nevertheless, as visualized in Figure \ref{fig:probing}, the flow patterns in the diagrams indicate that the distribution across languages does not exhibit any strong separability. The color continuity from the language sources through the personality type nodes toward the final classifier outputs supports the conclusion that the classifier's decisions are not dominated by language-specific signals.

Overall, the results demonstrate embedding neutrality between PIGA and TOGA and support the validity of our experimental setting. However, since our evaluation is \textbf{limited} to benchmark datasets, it does not fully reflect real-world multilingual distributions. We therefore acknowledge that validating the method on naturally annotated multilingual corpora remains an important direction for \textbf{future research}.

\subsection{CLAD Analysis}
In the CLAD subsection, we evaluate the proposed algorithm by analysing its impact on downstream task performance across multilingual evaluation datasets, where CLAD is designed to align performance consistently across languages and can be jointly trained with the target language data. In addition, we conduct an ablation study to isolate the contribution of each component within CLAD, clarifying their individual and combined effects. We further examine computational complexity and we report benchmarking results against standard training strategies to demonstrate its effectiveness in cross-lingual generalization. 

\subsubsection{Performance Implications}

Overall, the results demonstrate that CLAD consistently outperforms both the zero-shot cross-lingual model and the baseline model trained with weighted BCE. 

On the Essays dataset, CLAD achieves a balanced accuracy (BA) of 0.6204 with TOGA augmentation, corresponding to an improvement of +0.0641, and 0.7305 on the Kaggle dataset, yielding a +0.0789 gain. When applying PIGA augmentation, CLAD further improves to a BA of 0.6332 on the Essays dataset (+0.0573) and 0.7448 on the Kaggle dataset (+0.0968). As shown in Figure~\ref{fig:stats}, these improvements are statistically significant at $p < 0.05$, as confirmed by the McNemar test.

\begin{figure}
    \centering
    \subfigure[Essays Dataset]{
        \includegraphics[width=0.45\linewidth]{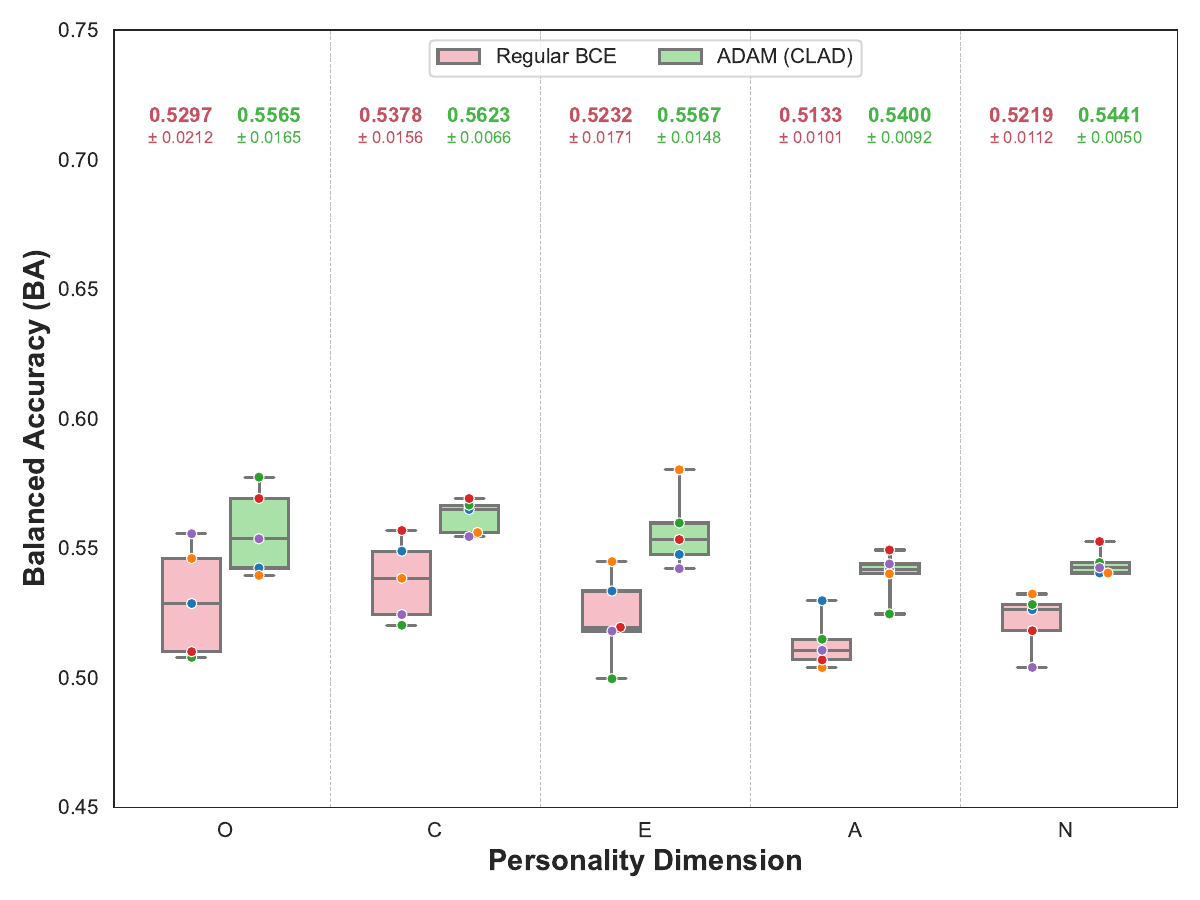}
        \label{fig:generalise-essays}
    }
    \subfigure[Kaggle Dataset]{
        \includegraphics[width=0.45\linewidth]{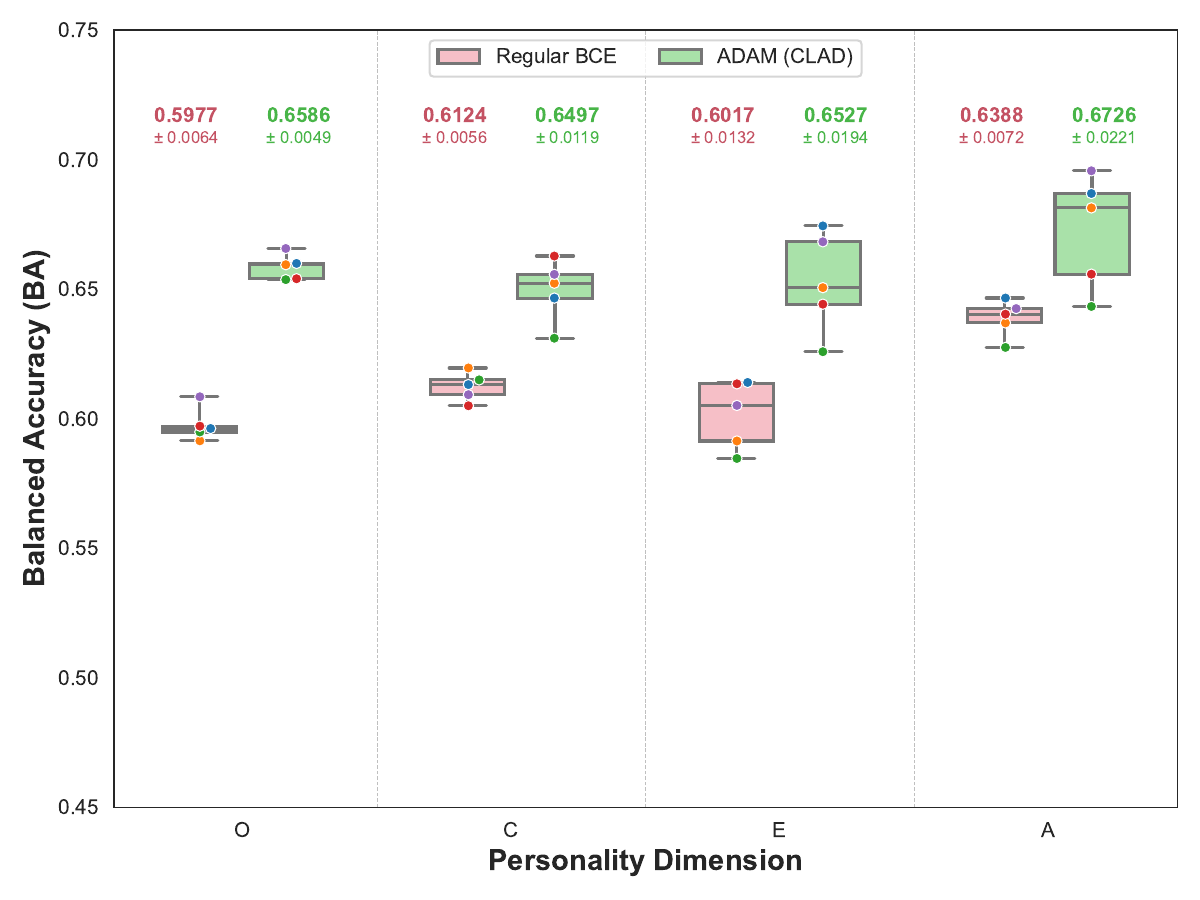}
        \label{fig:generalise-kaggle}
    }
\caption{Generalizable zero-shot cross-lingual evaluation of the model trained with weighted BCE and ADAM (CLAD) using PIGA augmentation. The model is trained on four languages and evaluated on a fifth (unseen) language across different language combinations. ADAM (CLAD) demonstrates statistically better generalization performance.}
    \label{fig:generalise}
\end{figure}

In contrast, the zero-shot cross-lingual model is trained solely on English data and does not have access to other languages during training. Nevertheless, it exhibits a certain degree of generalizability, performing above chance level despite the absence of multilingual supervision. Motivated by this observation, we further analyze the generalization capability of the proposed CLAD framework in comparison to the zero-shot model. Figure~\ref{fig:generalise} presents the results when the model is trained on four languages and evaluated on a held-out fifth language. The cross-lingual transfer performance of the proposed ADAM (CLAD) further substantiates its generalizability, demonstrating strong zero-shot adaptation capability to previously unseen languages, with improvements of 2.7\% on the Essays dataset and 4.6\% on the Kaggle dataset in average.

\subsubsection{Ablation Study}

\begin{figure*}
    \centering
    \subfigure[English (\texttt{en})]{
        \includegraphics[width=0.18\textwidth]{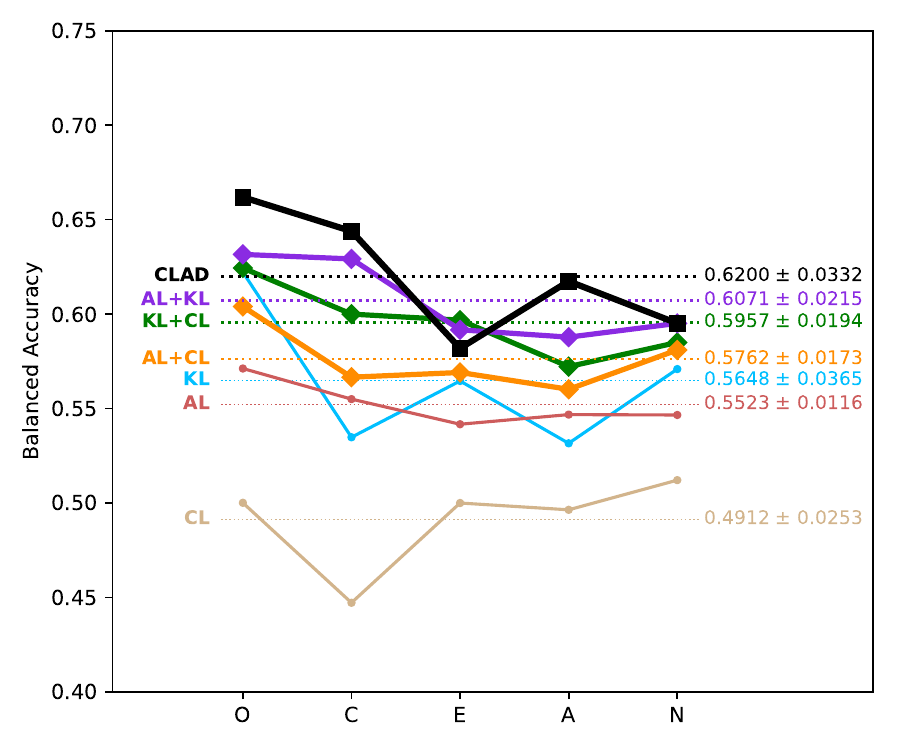}
        \label{fig:ea_en}
    }
    \subfigure[French (\texttt{fr})]{
        \includegraphics[width=0.18\textwidth]{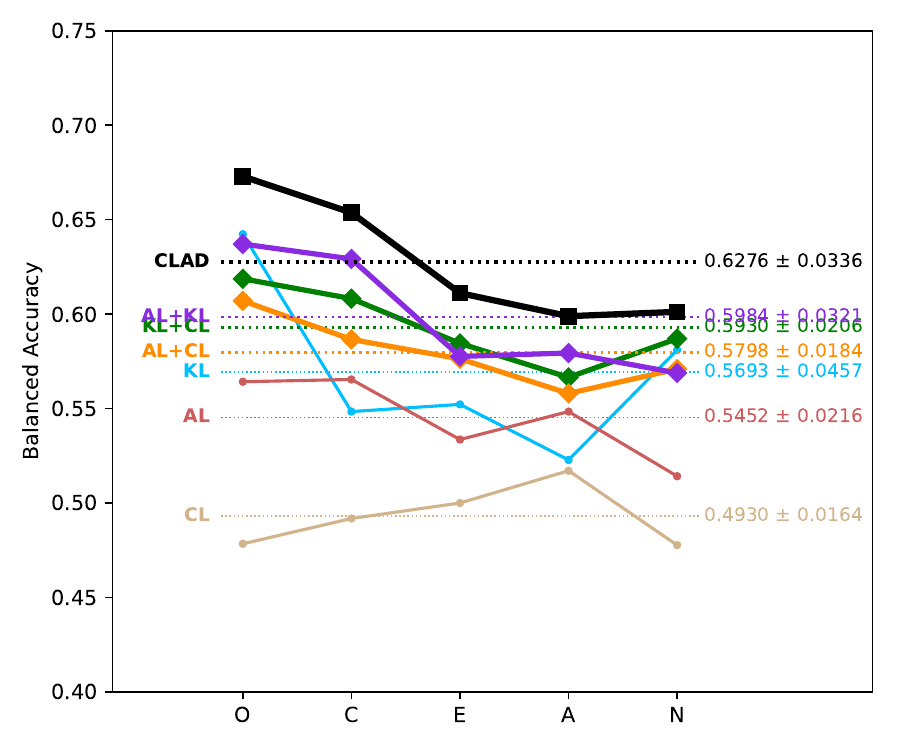}
        \label{fig:ea_fr}
    }
    \subfigure[Malay (\texttt{bm})]{
        \includegraphics[width=0.18\textwidth]{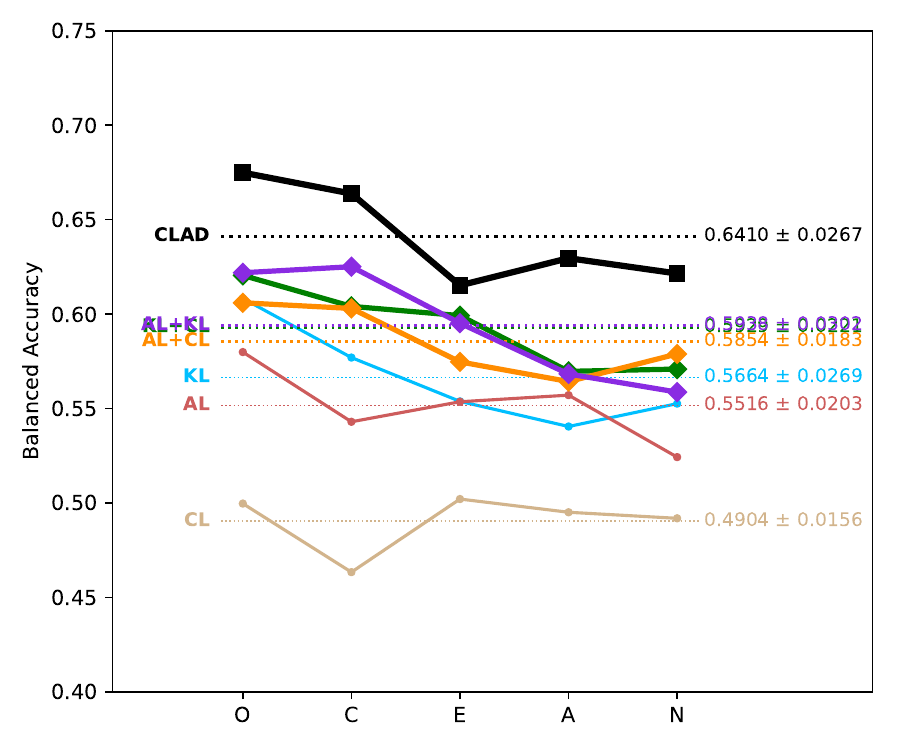}
        \label{fig:ea_bm}
    }
    \subfigure[Japanese (\texttt{jp})]{
        \includegraphics[width=0.18\textwidth]{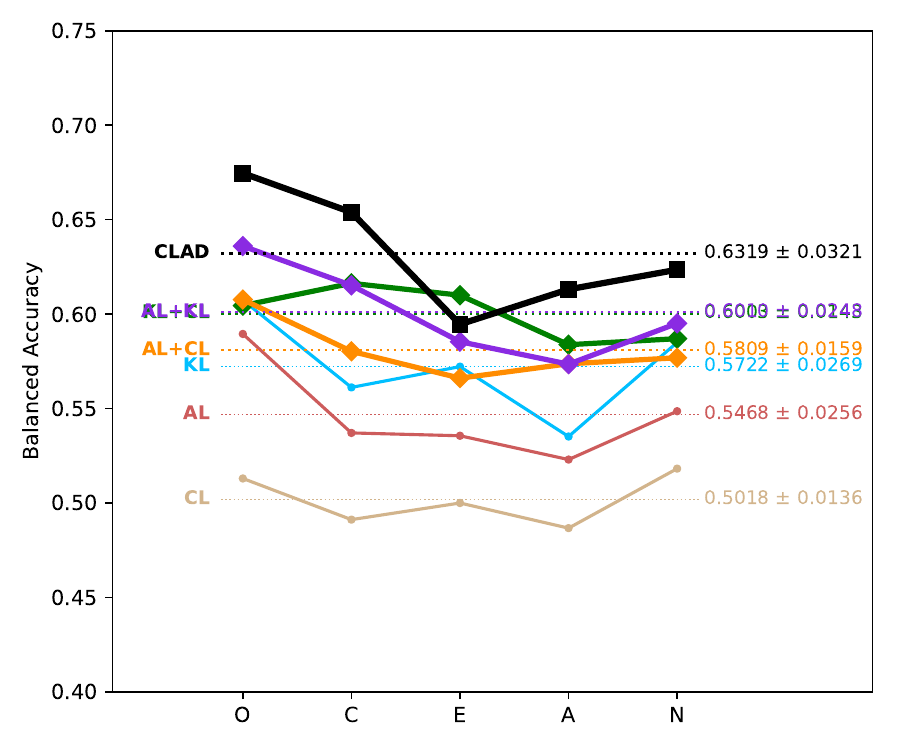}
        \label{fig:ea_jp}
    }
    \subfigure[Chinese (\texttt{zh})]{
        \includegraphics[width=0.18\textwidth]{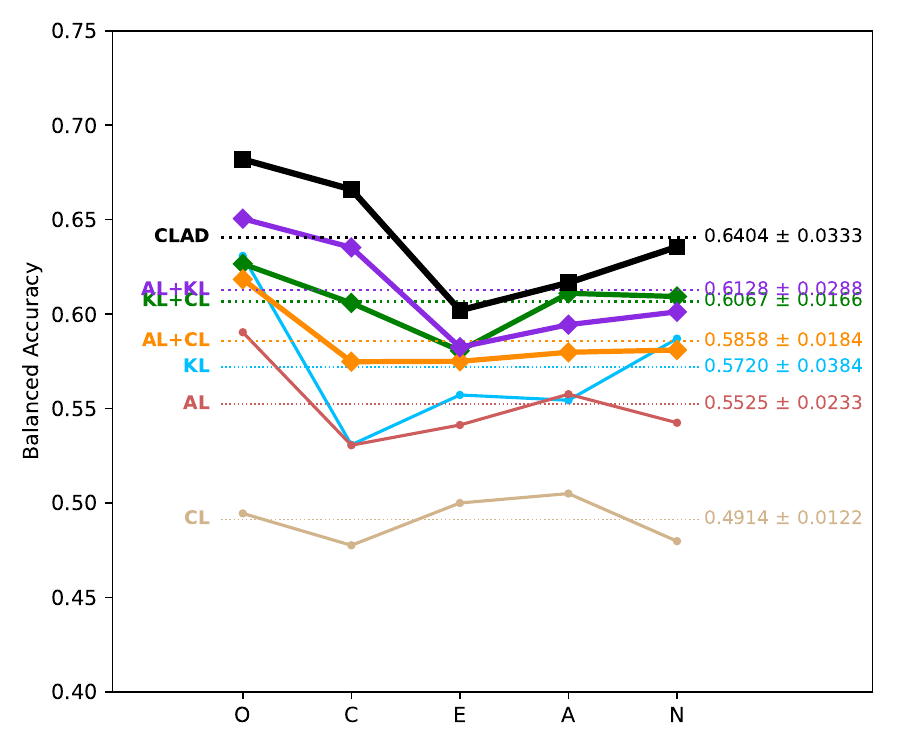}
        \label{fig:ea_zh}
    }
    \caption{
        Balanced Accuracy comparison of CLAD and its ablation components across five languages (\texttt{en}, \texttt{fr}, \texttt{bm}, \texttt{jp}, \texttt{zh}) on the Essays dataset. The full CLAD model consistently outperforms all ablations—(i) KL only, (ii) AL only, (iii) CL only, (iv) AL + CL, (v) AL + KL, (vi) CL + KL—demonstrating the effectiveness of integrating all components for multilingual personality recognition.}
    \label{fig:essays-ablation}
\end{figure*}

\begin{figure*}
    \centering
    \subfigure[English (\texttt{en})]{
        \includegraphics[width=0.18\textwidth]{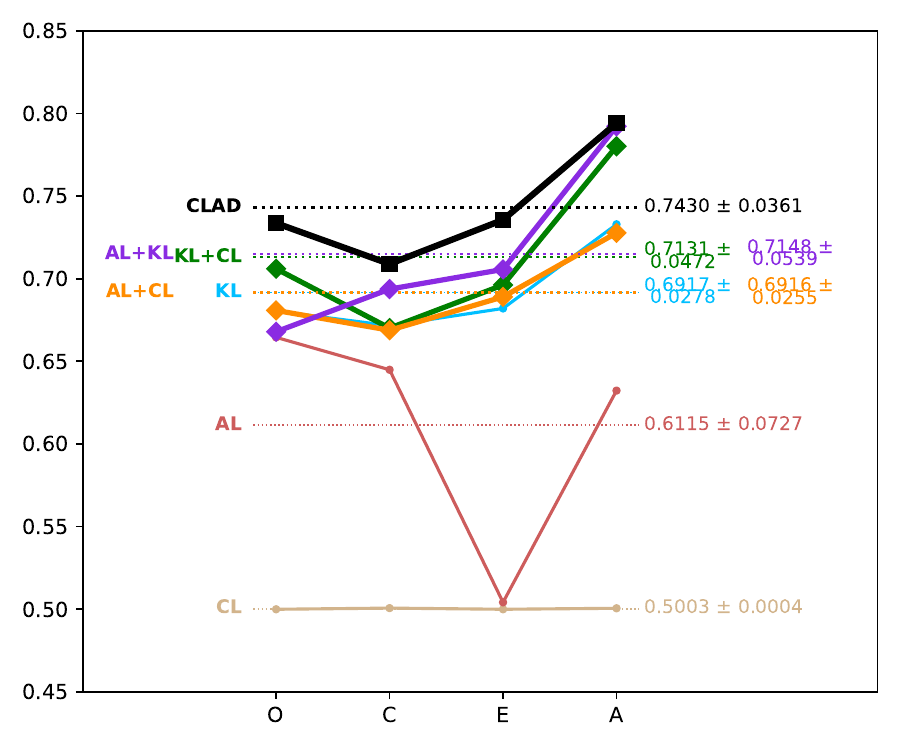}
        \label{fig:ka_en}
    }
    \subfigure[French (\texttt{fr})]{
        \includegraphics[width=0.18\textwidth]{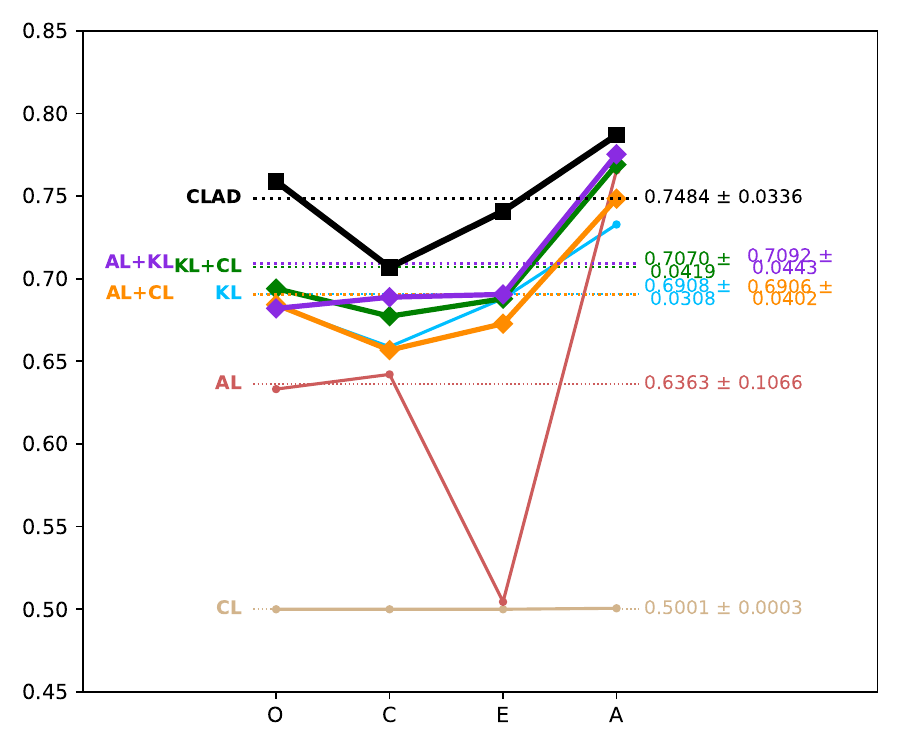}
        \label{fig:ka_fr}
    }
    \subfigure[Malay (\texttt{bm})]{
        \includegraphics[width=0.18\textwidth]{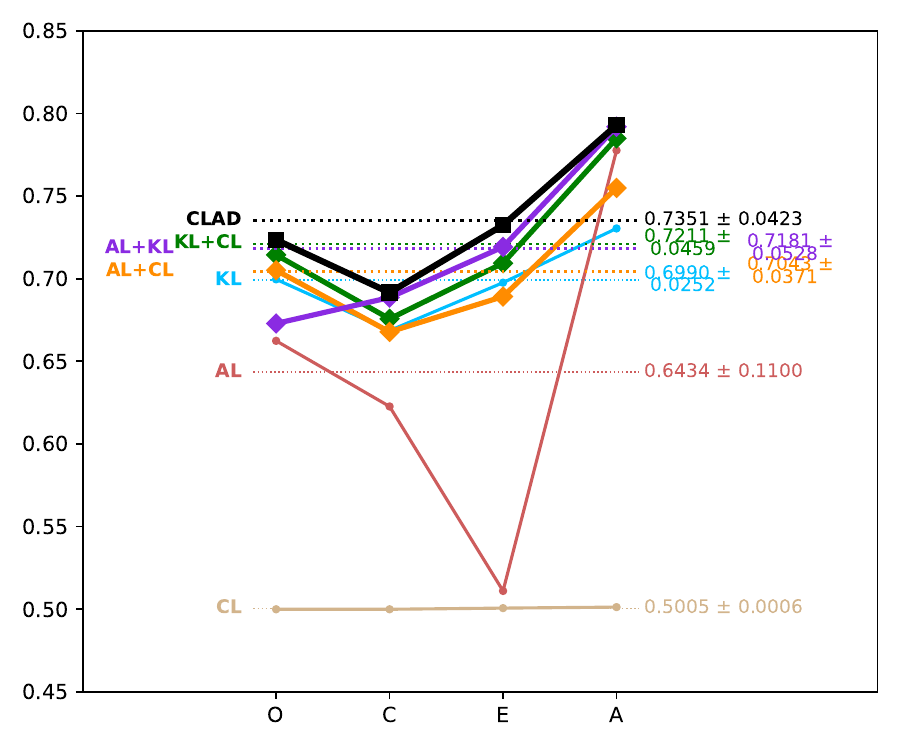}
        \label{fig:ka_bm}
    }
    \subfigure[Japanese (\texttt{jp})]{
        \includegraphics[width=0.18\textwidth]{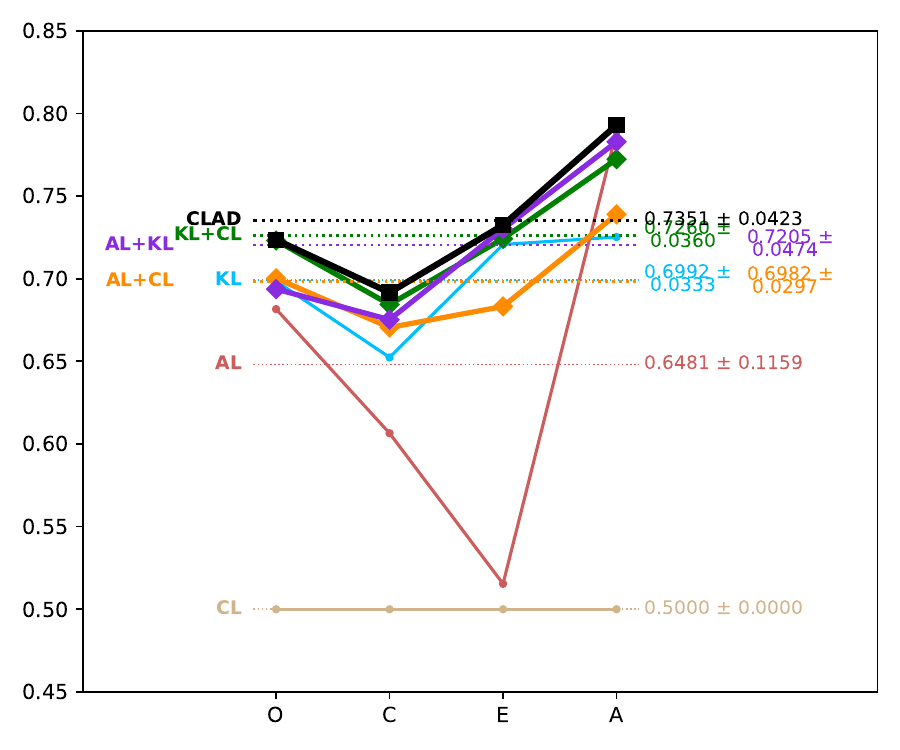}
        \label{fig:ka_jp}
    }
    \subfigure[Chinese (\texttt{zh})]{
        \includegraphics[width=0.18\textwidth]{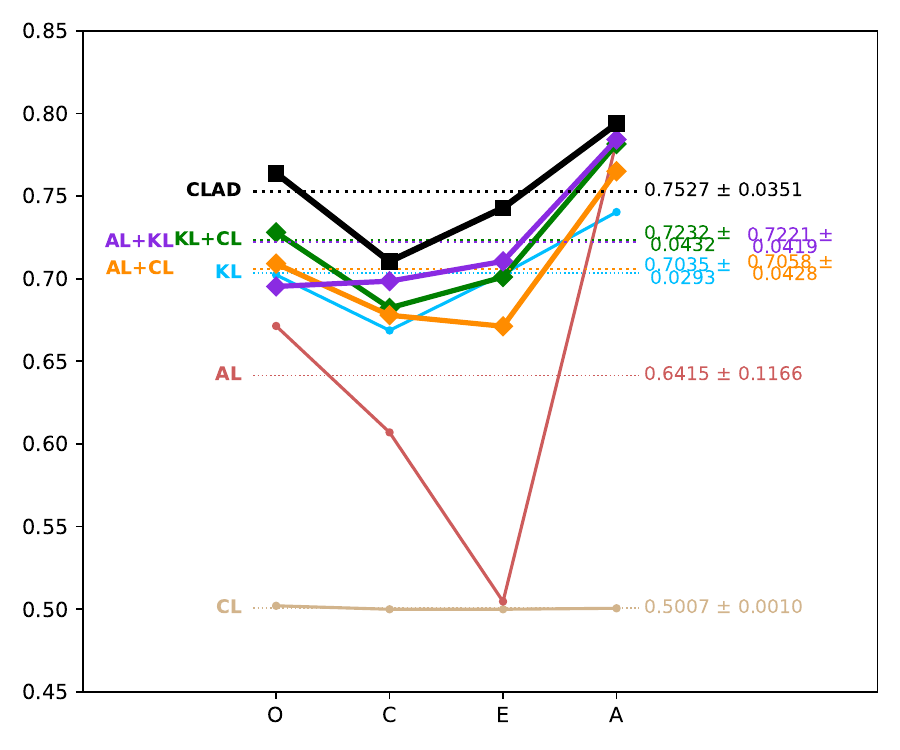}
        \label{fig:ka_zh}
    }
    \caption{
        Balanced Accuracy comparison of CLAD and its ablation components across five languages (\texttt{en}, \texttt{fr}, \texttt{bm}, \texttt{jp}, \texttt{zh}) on the Kaggle dataset. The full CLAD model consistently outperforms all ablations—(i) KL only, (ii) AL only, (iii) CL only, (iv) AL + CL, (v) AL + KL, (vi) CL + KL—demonstrating the effectiveness of integrating all components for multilingual personality recognition.}
    \label{fig:kaggle-ablation}
\end{figure*}

To further analyze the contribution of each CLAD component, we conducted an ablation study, as shown in Fig. \ref{fig:essays-ablation} and Fig. \ref{fig:kaggle-ablation}, for the Essays and Kaggle datasets, respectively. The study evaluates the impact of each individual module within CLAD. Each module corresponds to a specific language aspect aligned with a personality dimension, depending on the dataset. Notably, the Essays dataset is relatively balanced across personality traits, whereas the MBTI (Kaggle) dataset is more imbalanced—particularly for the Openness (O), Conscientiousness (C), and Extraversion (E) dimensions, which resulting a slightly different trend in the ablation study.

Since Contrastive Loss (CL) intended merely to align embeddings across different languages and is not designed for classification tasks, we observed that its performance is the worst, regardless of the dimension or dataset. This results in a random-guessing performance, with accuracy around 0.5 for accuracy.

Later, we observed that using only the KL divergence loss yields the best performance among the individual modules. This is because the KL module is designed to transfer language-specific embeddings from the teacher to the student model. However, when we conducted experiments by matching only English-to-English data (i.e., without involving augmentation), the student model was only able to learn English embeddings from the teacher model. The improvements for other languages were very limited—until we began matching embeddings across languages. This cross-lingual matching enabled the generated logits to represent the underlying personality rather than the original content semantics that the pre-trained student model was initially designed to capture.

The AL module is designed to automatically weight the contribution of each backpropagation signal during training, in order to reduce noise from certain dimensions or languages. In some cases, it does show better performance than the KL module alone. However, on average, it is less effective. This is because the AL module tends to be sensitive to easier dimensions or languages. This behavior is particularly evident in the `E' dimension of the Kaggle dataset, where it appears to be ignored during training, while the model focuses more on the `A' dimension, which is more balanced and easier to learn. In fact, performance on the `A' dimension can sometimes match that of the full CLAD model (e.g., for Japanese in the Kaggle dataset).

In both datasets, the AL+CL combination performs less effectively than the AL+KL and KL+CL combinations. Given the previously discussed contribution of the KL module, it is evident that KL is the primary driver of high performance, thanks to the strong supervision provided by the teacher model. Meanwhile, both AL and CL still outperform the AL-only model. This suggests that cross-lingual learning plays an important role in compensating for the limitations of the AL module, which tends to focus only on easier dimensions or languages. Although the CL module is not originally designed for this task, it contributes to improved performance by enhancing the model’s ability to generalize across languages.

For both AL+KL and KL+CL combinations, we observed nearly identical performance, confirming that the KL module is the main driver of improvement. On average, incorporating AL tends to yield more benefit than CL, as AL helps reduce noise during training, and its limitations—such as overlooking certain dimensions—are largely compensated by the KL component. While CL remains effective when combined with KL by facilitating cross-lingual embedding alignment, a key \textbf{limitation} of the AL mechanism is that its attention strategy may inadvertently prioritize easier samples to suppress noise from LLM-generated data. This behavior, however, may reduce the emphasis on harder or minority-class samples, which are important for addressing class imbalance. Mitigating this trade-off between noise reduction and hard-sample learning is a promising direction for \textbf{future work}. Overall, these observations reinforce our confidence in the effectiveness of the full CLAD framework.

Generally, the full CLAD model consistently outperforms all ablation variants across all languages and datasets. One exception is the AL module in the `A' dimension in the Kaggle dataset, which performs comparably. However, even in this case, the CLAD model demonstrates superior average performance. This highlights the effectiveness of integrating all components, leveraging their strengths while compensating for their individual limitations, for multilingual personality recognition.

\subsubsection{Computational Complexity}
In terms of computational complexity, CLAD introduces only minimal overhead beyond the standard training cost of the student model. Beyond the regular $\mathcal{O}(\theta_{\text{student}})$ complexity, CLAD adds approximately $\mathcal{O}(\theta_{\text{projecting}}) + \mathcal{O}(|s|) + \mathcal{O}(|s|\cdot|d|)$, where $\theta_{\text{projecting}}$ corresponds to a small fully connected projection network used for aligning student and teacher embeddings. Compared to the transformer backbone of the student model, the cost of this projection network is negligible. Additional weighting overhead is also minimal, assuming predefined matching rules and parallel GPU execution. The teacher model's forward pass ($\theta_{\text{teacher}}$) is not required at every epoch, as its outputs can be precomputed and reused. Since $|s|$ and $|d|$ are small (typically 4 or 5), these extra computations are minimal (compare to transformer model). Meanwhile, CLAD consistently delivers substantial performance gains across all dimensions compared to weighted BCE training (both trained individually for each language or jointly in a multilingual setting). Moreover, CLAD exhibits faster convergence (in PIGA dataset), typically reaching optimal performance within 18 epochs, whereas weighted BCE training requires 52 epochs in the multilingual setup or an average of 31.2 epochs across five separately trained languages, yet still fails to achieve comparable results.

Moreover, by consolidating all languages into a single multilingual student model (the CLAD approach), the overall inference complexity is reduced from \(\mathcal{O}(|s|*\theta_{\text{student}})\) (separate models for each language) to \(\mathcal{O}(\theta_{\text{student}})\), since only one model is required for training and inference across all languages. This design not only enhances computational practicality but also allows the model to leverage the shared cross-lingual representations learned through the multilingual transformer backbone.

\subsubsection{Benchmarking}

Table \ref{tab:essays-bench} and Table \ref{tab:kaggle-bench} present a comprehensive comparison of multiple encoder backbones and the ADAM (CLAD) approach on both the Essays and Kaggle datasets. As shown in Table~\ref{tab:essays-bench}, the proposed ADAM (CLAD) framework consistently outperforms strong multilingual baselines, including \textit{multilingual-e5-large-instruct} \cite{multilingual-e5-large-instruct}, \textit{Qwen3-Embedding-0.6B} \cite{Qwen3-Embedding-0.6B}, \textit{embeddinggemma-300m} \cite{embeddinggemma-300m}, \textit{LaBSE} \cite{LaBSE}, and \textit{bge-m3} \cite{bge-m3}, across all OCEAN traits in terms of both BA and F1-score. The proposed framework is further adapted to the aforementioned encoder backbones with consistent performance improvements observed across different model variants.

\begin{table*}[t]
  \centering
  \scriptsize
  \caption{Benchmark comparison of different encoder backbone models trained with weighted binary cross-entropy (BCE) loss and the proposed ADAM (CLAD) model on the Essays dataset.}
\begin{tabular}{|l|r|r|r|r|r|r|r|r|r|r|r|r|}
\hline
\multicolumn{1}{|c|}{\multirow{2}{*}{\textbf{Model/Approach}}} & \multicolumn{2}{c|}{\textbf{O}} & \multicolumn{2}{c|}{\textbf{C}} & \multicolumn{2}{c|}{\textbf{E}} & \multicolumn{2}{c|}{\textbf{A}} & \multicolumn{2}{c|}{\textbf{N}} & \multicolumn{2}{c|}{\textbf{Average}}\\
\cline{2-13}  & \multicolumn{1}{l|}{BA} & \multicolumn{1}{l|}{F1} & \multicolumn{1}{l|}{BA} & \multicolumn{1}{l|}{F1} & \multicolumn{1}{l|}{BA} & \multicolumn{1}{l|}{F1} & \multicolumn{1}{l|}{BA} & \multicolumn{1}{l|}{F1} & \multicolumn{1}{l|}{BA} & \multicolumn{1}{l|}{F1} & \multicolumn{1}{l|}{BA} & \multicolumn{1}{l|}{F1}\\
\hline
\textit{multilingual-e5-large-instruct}\cite{multilingual-e5-large-instruct} & 0.5977 & 0.5617 & 0.5160 & 0.5100 & 0.5516 & 0.5783 & 0.5209 & 0.5612 & 0.5231 & 0.5542 & 0.5419 & 0.5531\\
\hline
\textit{Qwen3-Embedding-0.6B}\cite{Qwen3-Embedding-0.6B} & 0.5842 & 0.5559 & 0.5170 & 0.5179 & 0.5621 & 0.5946 & 0.5225 & 0.5658 & 0.5291 & 0.5710 & 0.5430 & 0.5610\\
\hline
\textit{embeddinggemma-300m} \cite{embeddinggemma-300m}& 0.5965 & 0.5729 & 0.5156 & 0.5111 & 0.5426 & 0.5768 & 0.5163 & 0.5582 & 0.5247 & 0.5552 & 0.5391 & 0.5548\\
\hline
\textit{LaBSE} \cite{LaBSE} & 0.5906 & 0.5636 & 0.5397 & 0.5392 & 0.5620 & 0.5955 & 0.5271 & 0.5598 & 0.5227 & 0.5669 & 0.5484 & 0.5650\\
\hline
\textit{bge-m3} \cite{bge-m3}& 0.6221 & 0.5922 & 0.5444 & 0.5372 & 0.5577 & 0.5887 & 0.5631 & 0.5952 & 0.5648 & 0.5921 & 0.5704 & 0.5811\\
\hline
\textbf{ADAM (CLAD)} & \textbf{0.6732} & \textbf{0.6526} & \textbf{0.6563} & \textbf{0.6593} & \textbf{0.6010} & \textbf{0.6263} & \textbf{0.6151} & \textbf{0.6804} & \textbf{0.6154} & \textbf{0.5989} & \textbf{0.6322} & \textbf{0.6435}\\
\hline
\end{tabular}
  \label{tab:essays-bench}
  
\end{table*}

\begin{table*}[t]
  \centering
  \scriptsize
  \caption{Benchmark comparison of different encoder backbone models trained with weighted binary cross-entropy (BCE) loss and the proposed ADAM (CLAD) model on the Kaggle dataset.}
\begin{tabular}{|l|r|r|r|r|r|r|r|r|r|r|}
\hline
\multicolumn{1}{|c|}{\multirow{2}{*}{\textbf{Model/Approach}}} & \multicolumn{2}{c|}{\textbf{O}} & \multicolumn{2}{c|}{\textbf{C}} & \multicolumn{2}{c|}{\textbf{E}} & \multicolumn{2}{c|}{\textbf{A}} & \multicolumn{2}{c|}{\textbf{Average}} \\
\cline{2-11}  & \multicolumn{1}{l|}{BA} & \multicolumn{1}{l|}{F1} & \multicolumn{1}{l|}{BA} & \multicolumn{1}{l|}{F1} & \multicolumn{1}{l|}{BA} & \multicolumn{1}{l|}{F1} & \multicolumn{1}{l|}{BA} & \multicolumn{1}{l|}{F1} & \multicolumn{1}{l|}{BA} & \multicolumn{1}{l|}{F1} \\
\hline
\textit{multilingual-e5-large-instruct} \cite{multilingual-e5-large-instruct} & 0.6614 & 0.8496 & 0.6445 & 0.5700 & 0.6785 & 0.5194 & 0.7496 & 0.7600 & 0.6835 & 0.6748 \\
\hline
\textit{Qwen3-Embedding-0.6B} \cite{Qwen3-Embedding-0.6B}  & 0.6390 & 0.8470 & 0.6388 & 0.5632 & 0.6622 & 0.4864 & 0.7526 & 0.7634 & 0.6732 & 0.6650 \\
\hline
\textit{embeddinggemma-300m} & 0.6450 & 0.8466 & 0.6492 & 0.5773 & 0.6702 & 0.5029 & 0.7249 & 0.7366 & 0.6723 & 0.6659 \\
\hline
\textit{LaBSE} \cite{LaBSE} & 0.6785 & 0.8650 & 0.6563 & 0.5830 & 0.6696 & 0.5013 & 0.7384 & 0.7493 & 0.6857 & 0.6746 \\
\hline
\textit{bge-m3} \cite{bge-m3}  & 0.6853 & 0.8653 & 0.6649 & 0.5964 & 0.6863 & 0.5353 & 0.7575 & 0.7670 & 0.6985 & 0.6910 \\
\hline
\textbf{ADAM (CLAD)} & \textbf{0.7468} & \textbf{0.8322} & \textbf{0.7011} & \textbf{0.6500} & \textbf{0.7405} & \textbf{0.6085} & \textbf{0.7908} & \textbf{0.8167} & \textbf{0.7448} & \textbf{0.7269} \\
\hline
\end{tabular}%
\label{tab:kaggle-bench}  
\end{table*}

Overall, ADAM (CLAD) achieves the highest BA and F1-score on both the Essays and Kaggle datasets. Although in certain dimensions the F1-score on the Kaggle dataset is not the highest due to class imbalance, the results further demonstrate the effectiveness of CLAD in mitigating imbalance effects and improving performance under challenging settings~\cite{Tan2025aflps}. The improvement can be attributed to the knowledge distillation mechanism, which enables effective knowledge transfer from the teacher model, and the attention-based alignment strategy, which facilitates balanced contribution across languages and enhances cross-lingual representation learning.

\section{Conclusion}
In this paper, we present ADAM, a framework for multilingual personality recognition. We first propose PIGA to leverage the generative capabilities of large language models (LLMs) to augment personality datasets with multilingual data. This generated data is then analyzed using our linguistic analysis pipeline and a probing classifier to validate the plausibility of the assigned personality labels. Later, we introduce CLAD, a cross-lingual augmentation framework that employs a teacher-student architecture with an attention mechanism to reduce noise introduced during augmentation. CLAD also incorporates a contrastive loss function to effectively capture personality-related features across different languages. Experimental results show that the proposed CLAD algorithm, when combined with PIGA augmentation, achieves average balanced accuracy (BA) scores of 0.6332 on the Essays dataset and 0.7448 on the Kaggle dataset across five languages (\texttt{en}, \texttt{fr}, \texttt{bm}, \texttt{jp}, \texttt{zh}). These results represent improvements of and 5.7\% and 9.7\% over the baseline model, respectively. Future work could investigate relaxing the binary trait assumption by modeling personality dimensions in continuous or ordinal spaces, which may reduce discretization effects and better capture trait variability in multilingual settings.

\section*{Acknowledgment}
This research was funded by the Universiti Tunku Abdul Rahman Research Fund (IPSR/RMC/UTARRF/2021-C1/K03). The authors gratefully acknowledge the support of Université Sorbonne Paris Nord for providing technical and research assistance through access to Grid5000, which facilitated the computational resources used in this study. The authors also thank the Advanced Artificial Intelligence Research Center at the Kanagawa Institute of Technology for supporting OpenAI's model inferences.

\bibliographystyle{IEEEtran}
\bibliography{references}

\newpage

\section*{Author Biographies}

\begin{IEEEbiography}[{\includegraphics[width=1in,height=1.25in,clip,keepaspectratio]{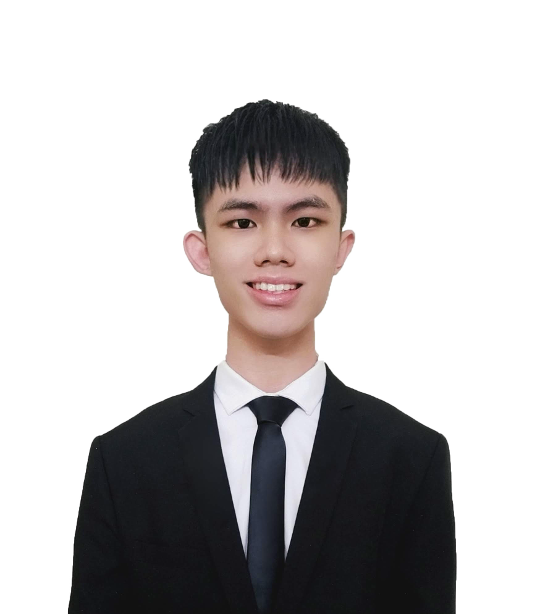}}]{Jing Jie Tan}received the Bachelor of Computer Science (Hons) degree and is completing his Ph.D. in Engineering, specializing in machine learning, at Universiti Tunku Abdul Rahman (UTAR). He is a Research Fellow at the National University of Singapore (NUS). His research interests include deep learning, natural language processing, computer vision, large language models, vision transformers, and cross-disciplinary research, with a focus on translating research into practical applications.
\end{IEEEbiography}

\begin{IEEEbiography}[{\includegraphics[width=1in,height=1.25in,clip,keepaspectratio]{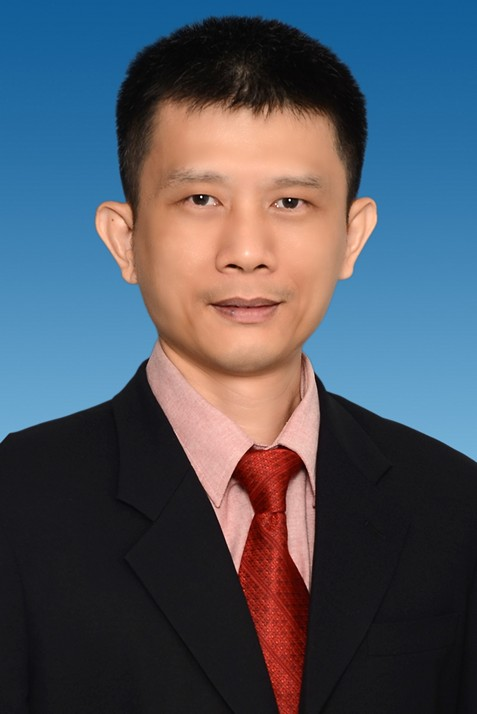}}]{Ban-Hoe Kwan} received the Bachelor of Engineering (Electrical), Master of Engineering Science, and Ph.D. degrees in Engineering from the University of Malaya (UM). He is currently an Associate Professor at Universiti Tunku Abdul Rahman (UTAR). His research interests include image processing, artificial intelligence, medical signal processing, the Internet of Things, and robotics.
\end{IEEEbiography}

\begin{IEEEbiography}[{\includegraphics[width=1in,height=1.25in,clip,keepaspectratio]{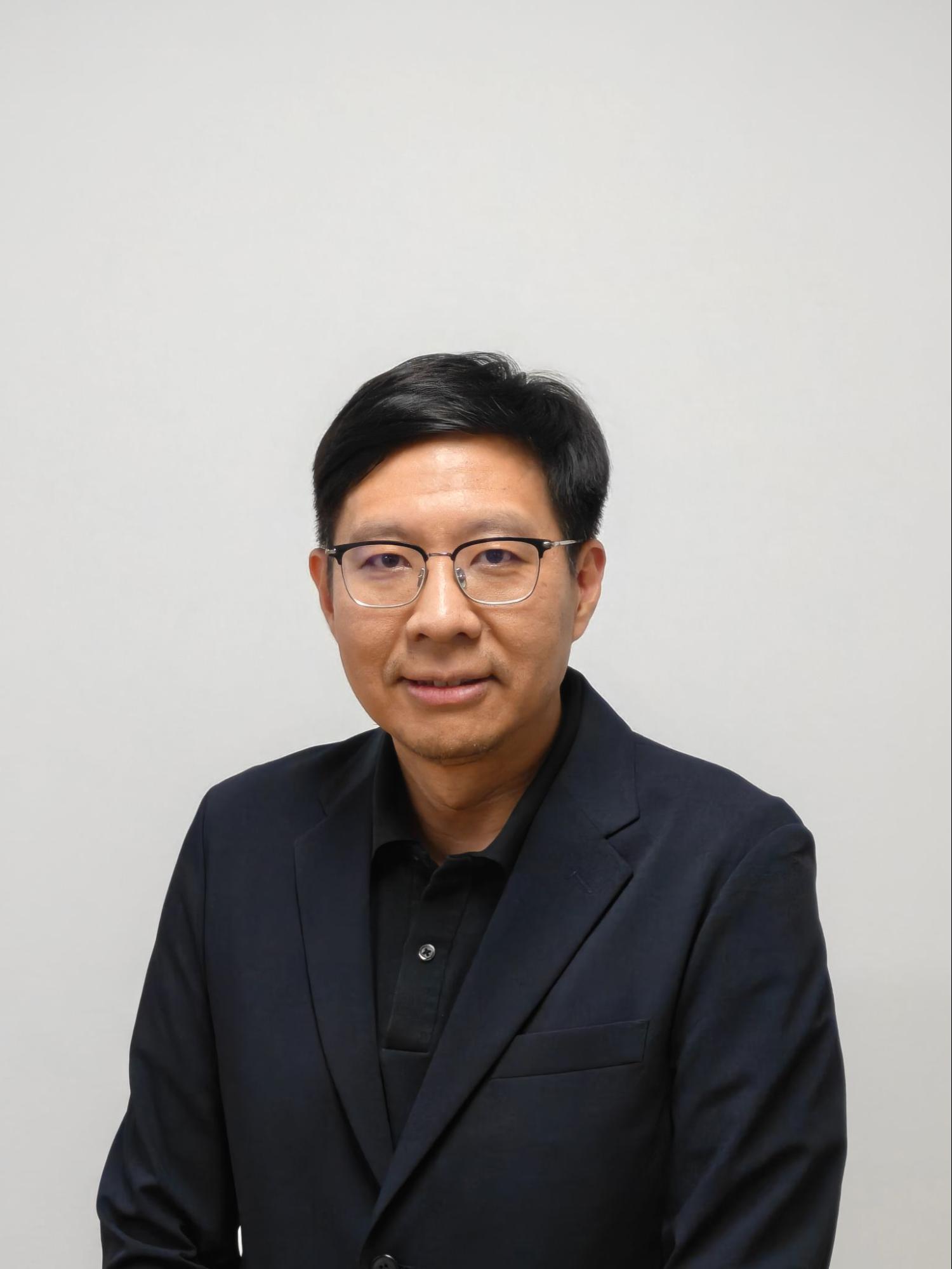}}]{Danny Ng Wee Kiat} (Senior Member, IEEE) received the Ph.D. degree in Engineering and is a registered Professional Engineer with the Board of Engineers Malaysia. He is currently an Assistant Professor at Universiti Tunku Abdul Rahman (UTAR). His research focuses on robotics and artificial intelligence, particularly AI-driven robotics, generative AI, and autonomous AI agents for industrial and enterprise applications. 

In addition to his academic role, he serves as the Chief Executive Officer and Chief Technology Officer of Netizen Robotics and as the Technical Director of Netizen Experience, where he leads initiatives in advanced robotics and digital transformation.
\end{IEEEbiography}

\begin{IEEEbiography}[{\includegraphics[width=1in,height=1.25in,clip,keepaspectratio]{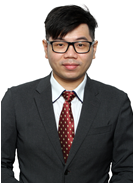}}]{Yan Chai Hum} (Senior Member, IEEE) received the Ph.D. degree in Engineering with specialization in artificial intelligence from Universiti Teknologi Malaysia. He is currently an Associate Professor with the Department of Mechatronics and Biomedical Engineering, Lee Kong Chian Faculty of Engineering and Science, Universiti Tunku Abdul Rahman (UTAR). His research interests include artificial intelligence in healthcare, biomedical imaging, computer vision, Internet of Things systems, and intelligent sensing technologies. His recent work focuses on AI-driven diagnostic systems, multimodal medical data analysis, and autonomous sensing platforms. He is also the founder of Promptiq Enterprise, an AI consultancy specializing in generative AI and intelligent systems.
\end{IEEEbiography}

\begin{IEEEbiography}[{\includegraphics[width=1in,height=1.25in,clip,keepaspectratio]{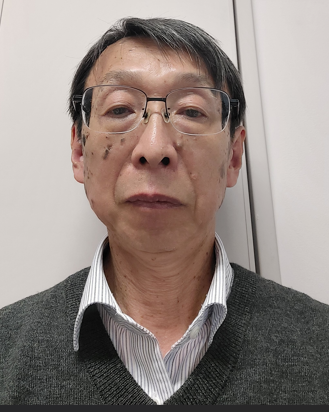}}]{Noriyuki Kawarazaki} received the Ph.D. degree in Engineering from Kyushu University. He is currently a Professor and Department Chair of Information Systems at Kanagawa Institute of Technology, Japan. His research interests include human-robot interaction, image processing, and artificial intelligence in robotics.
\end{IEEEbiography}

\begin{IEEEbiography}[{\includegraphics[width=1in,height=1.25in,clip,keepaspectratio]{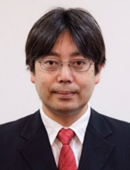}}]{Kosuke Takano} received the B.A. degree in Environment and Information Studies and the M.A. and Ph.D. degrees in Media and Governance from Keio University, Japan. He is currently a Professor in the Department of Information and Computer Sciences at Kanagawa Institute of Technology. His research interests include emotional AI, multimodal affective computing, data management systems, and AI-driven educational systems.
\end{IEEEbiography}

\begin{IEEEbiography}[{\includegraphics[width=1in,height=1.25in,clip,keepaspectratio]{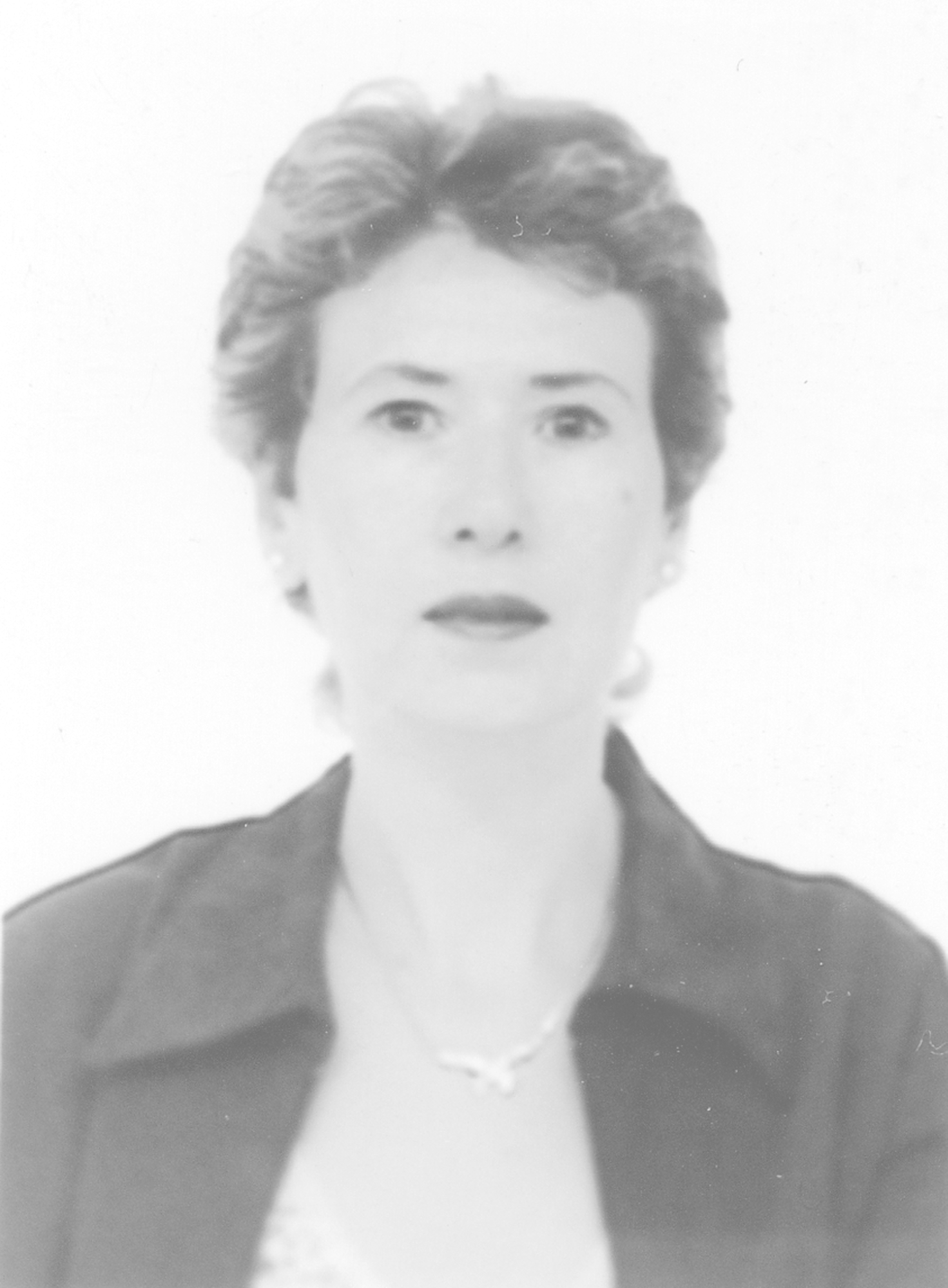}}]{Anissa Mokraoui} received the State Engineering degree in electrical engineering from the National School of Telecommunications, Algeria, in 1989, the M.S. degree in information technology in 1990, and the Ph.D. degree in 1994 from University Paris 11, Orsay, France. She is currently a Full Professor at Universit\'e Sorbonne Paris Nord (USPN), France, and Director of the L2TI Laboratory. Her research interests include source coding, joint source-channel decoding, robust transmission, MIMO-OFDM systems, computer vision, and few-shot object detection. She has authored over 150 publications and actively contributes to international conferences and journals as a reviewer and committee member.
\end{IEEEbiography}

\end{document}